%% file: main.tex
\documentclass[10pt,twocolumn,letterpaper]{article}
\usepackage[pagenumbers]{cvpr}
\usepackage{microtype}
\usepackage{graphicx}
\usepackage{multirow}
\usepackage{booktabs}
\usepackage{graphicx}
\usepackage{subcaption} 
\usepackage{tikz}
\usetikzlibrary{
    shapes.geometric,
    positioning,
    arrows.meta,
    shadows,
    fit,
    calc
}
\definecolor{cvprblue}{rgb}{0.21,0.49,0.74}
\usepackage[pagebackref,breaklinks,colorlinks,allcolors=cvprblue]{hyperref}

\title{RoadSceneBench: A Lightweight Benchmark for Mid-Level Road Scene Understanding}

\author{
Xiyan Liu$^*$ \quad
Han Wang \quad
Yuhu Wang \quad
Junjie Cai \quad
Zhe Cao \quad
Jianzhong Yang \quad
Zhen Lu \\
Baidu Inc.
}

\begin{document}

\twocolumn[{%
\renewcommand\twocolumn[1][]{#1}%
\maketitle
\begin{center}
    \includegraphics[width=\textwidth,trim={4cm 1.5cm 1.0cm 3.8cm},clip]{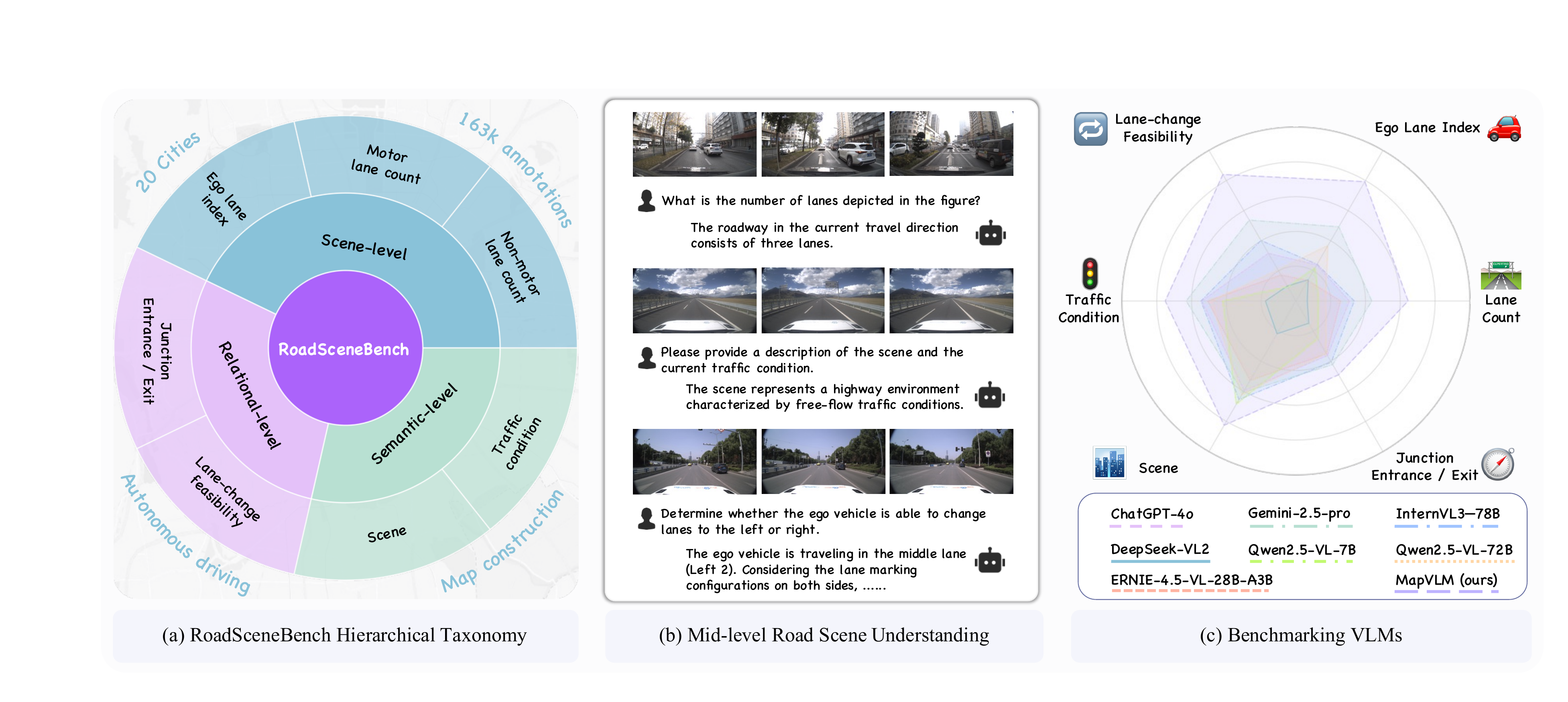}
    \vspace{-4mm}
    \captionof{figure}{
    \textbf{Overview of RoadSceneBench.}
    The benchmark spans Scene-, Relational-, and Semantic-level tasks with multi-frame reasoning. Furthermore, we benchmark various open- and closed-source models.
    }
    \label{fig:overview}
\end{center}
}]
\begingroup
\renewcommand\thefootnote{$*$}%
\footnotetext{Corresponding at: \texttt{liuxiyan@baidu.com}}%
\endgroup

\input{sec/0_abstract}    
\input{sec/1_intro}
\input{sec/2_related_work}

\input{sec/3_dataset}
\input{sec/4_method}

\input{sec/5_conclusion}
{
    \small
    \bibliographystyle{ieeenat_fullname}
    \bibliography{main}
}

\input{sec/X_suppl}

\end{document}

%% file: sec/0_abstract.tex
\begin{abstract}
Understanding mid-level road semantics, which capture the structural and contextual cues that link low-level perception to high-level planning, is essential for reliable autonomous driving and digital map construction. However, existing benchmarks primarily target perception tasks such as detection or segmentation, overlooking the reasoning capabilities required to infer road topology and dynamic scene structure. To address this gap, we present RoadSceneBench, a lightweight yet information-rich benchmark designed to evaluate and advance visual reasoning in complex road environments. Unlike large-scale perception datasets, RoadSceneBench emphasizes relational understanding and structural consistency, encouraging models to capture the underlying logic of real-world road scenes. Furthermore, to enhance reasoning reliability, we propose Hierarchical Relational Reward Propagation with Temporal Consistency (HRRP-T), a training framework for Vision-Language Models (VLMs) in which reward signals adaptively promote spatial coherence and semantic alignment throughout the reasoning process. This paradigm enables models to move beyond static recognition toward geometry-aware and temporally consistent reasoning. Extensive experiments demonstrate that our method achieves state-of-the-art performance across diverse road configurations. RoadSceneBench thus provides a compact yet powerful foundation for studying mid-level road semantics and fostering structure-aware autonomous perception. Our dataset is available at \url{https://github.com/XiyanLiu/RoadSceneBench}.

\end{abstract}

%% file: sec/1_intro.tex
\section{Introduction}
\label{sec:intro}

Automatic digit map construction and reliable autonomous driving rely not only on accurate perception but also on a deeper understanding of road scene semantics, which encompass the structural and contextual cues that bridge raw sensor data and high-level decision-making. While recent advances in detection and segmentation have significantly improved low-level visual perception, these tasks often fail to capture the relational and geometric reasoning abilities required to interpret how road elements interact. For instance, determining the feasibility of lane changes in adjacent lanes, identifying entrance or exit ramps, and inferring local traffic conditions are essential for predicting feasible maneuvers and maintaining map consistency. Such reasoning lies in the mid-level layer of scene understanding, connecting perception with planning, yet remains largely overlooked in existing datasets and benchmarks.

Current road-scene benchmarks, such as Cityscapes~\cite{cordts2016cityscapes}, BDD100K~\cite{yu2018bdd100k}, or nuScenes~\cite{caesar2020nuscenes}, primarily target dense annotation for pixel-wise segmentation, object detection, or 3D reconstruction. Although invaluable for perception, these datasets are less suited for studying structural consistency and semantic reasoning across road elements. In contrast, there is a growing need for compact, reasoning-oriented datasets that can effectively evaluate a model’s ability to infer and maintain the geometric logic of real-world road layouts without relying on heavy supervision. To address this gap, we present RoadSceneBench, a lightweight yet information-rich benchmark specifically designed for mid-level road semantic understanding. RoadSceneBench consists of $11,705$ front-view driving images that capture a diverse set of urban, suburban, and highway environments. Each image is annotated with interpretable mid-level attributes that reflect both structural (\textit{e.g.}, lane topology, entrance and exit ramps) and relational (\textit{e.g.}, whether adjacent lanes are crossable) aspects of road scenes, alongside contextual cues such as congestion states and ego-lane index. 

Additionally, to improve reasoning under such mid-level road semantics, we propose MapVLM, a framework based on Hierarchical Relational Reward Propagation with Temporal Consistency (HRRP-T). Rather than relying solely on static supervision, HRRP-T treats the inference process of vision-language models as a structured decision-making sequence, where each semantic attribute (such as lane count or ego-lane index) is generated under contextual dependencies. A reinforcement signal encourages predictions that maintain logical coherence within a frame and temporal consistency across short sequences. Specifically, HRRP-T integrates two complementary objectives: (1) structural reasoning, which enforces intra-frame topological validity (\textit{e.g.}, lane topology, cross-lane constraints), and (2) contrastive consistency, which aligns semantic reasoning across consecutive frames to encourage smooth evolution of spatial relations such as merging or congestion. Through self-critical reinforcement learning, the model optimizes these rewards without additional labels, unifying structure, semantics, and temporal context into a single adaptive reasoning paradigm that transforms VLMs from static predictors into geometry-aware reasoning agents for complex road scenes.

The key contributions are summarized as follows:
\begin{itemize}
\item We introduce RoadSceneBench, a benchmark dataset tailored for training and evaluating vision-language models (VLMs) on mid-level road scene understanding, a semantic layer that bridges the gap between low-level dense perception and high-level decision-making in autonomous driving and digital map construction.

\item We propose a unified VLM training paradigm that combines supervised fine-tuning with Hierarchical Relational Reward Propagation with Temporal Consistency (HRRP-T). This framework constrains the model from both structural reasoning and contrastive consistency perspectives, enabling geometry-aware and logically coherent understanding of road semantics.

\item Extensive experiments demonstrate the effectiveness of our approach and the practical utility of RoadSceneBench, underscoring its potential to advance research on mid-level road scene reasoning and promote more structured and interpretable road understanding.
\end{itemize}

%% file: sec/2_related_work.tex
\section{Related Work}
\label{sec:related}

\begin{table*}[t]
\centering
\caption{Comparison of autonomous driving scene benchmarks. RoadSceneBench uniquely provides \textbf{self-collected} data with \textbf{per-frame mid-level semantics}, bridging the gap between low-level perception and high-level reasoning.}
\vspace{-2mm}
\label{tab:benchmark_comparison}
\resizebox{\textwidth}{!}{%
\begin{tabular}{l|l|l|c|p{5.5cm}}
\toprule
\textbf{Category} & \textbf{Benchmark} & \textbf{Source} & \textbf{Scale} & \textbf{Primary Tasks} \\
\midrule
\multirow{2}{*}{Low-level 2D perception}
& Cityscapes~\cite{cordts2016cityscapes} & Self-collected (50 cities) & 5K fine + 20K coarse images & Semantic segmentation \\
& BDD100K~\cite{yu2018bdd100k} & Self-collected (US) & 100K images / videos & Segmentation, detection, tracking \\
\midrule
\multirow{2}{*}{Low-level 3D perception}
& nuScenes~\cite{caesar2020nuscenes} & Self-collected (Boston, Singapore) & 1000 scenes / $\sim$1.4M camera images & 3D detection, tracking, mapping \\
& Waymo Open Dataset~\cite{sun2020scalability} & Self-collected (US cities) & $\sim$1{,}150 segments (camera+LiDAR) & 3D detection, motion prediction \\
\midrule
\multirow{4}{*}{High-level VLM reasoning}
& nuScenes-QA~\cite{qian2024nuscenes} & Derived from nuScenes & $\sim$460K QA pairs & Visual question answering \\
& DriveLM~\cite{sima2024drivelm} & Derived from nuScenes & $\sim$360K QA pairs & Instruction following, graph reasoning \\
& VLADBench~\cite{li2025fine} & Derived from 12 public datasets & 12K questions / 5K scenes & Traffic knowledge, decision-making \\
& ROADWork~\cite{ghosh2025roadwork} & Self-collected / derived & $\sim$4.4K videos / $\sim$9.6K keyframes & Work-zone perception, risk analysis \\
\midrule
\textbf{Mid-level semantics}
& \textbf{RoadSceneBench (ours)} & \textbf{Self-collected (20 China cities)} & \textbf{2,341 clips (11,705 images), ~163K labels} & \textbf{Lane-topology and traffic-state understanding for autonomous driving and digital map-change reasoning.} \\
\bottomrule
\end{tabular}%
}
\vspace{-4mm}
\end{table*}

\subsection{VLMs and Structured Reasoning}
Large vision–language models (VLMs) trained on web-scale image–text pairs with instruction tuning exhibit strong zero/few-shot performance. Representative systems include CLIP~\cite{radford2021learning} for contrastive alignment, BLIP/BLIP-2~\cite{li2022blip,li2023blip}, Flamingo~\cite{alayrac2022flamingo}, InstructBLIP~\cite{dai2023instructblip}, and LLaVA~\cite{liu2023visual}, which couple high-capacity vision encoders with LLMs for general multimodal reasoning. Yet these models are primarily optimized for per-instance correctness and linguistic fluency, with limited guarantees on geometric consistency, discrete topological rules, or cross-task dependencies that are central to road-scene semantics.

Reinforcement learning further aligns models with sequence-level or human-preference objectives (\textit{e.g.}, self-critical sequence training~\cite{rennie2017self}, RLHF-style methods~\cite{christiano2017deep,rafailov2023direct}). In driving settings, VLM/LLM systems have been explored for generating explanations or high-level decisions based on perception outputs; however, most RL/preference optimization targets global behavior rather than enforcing fine-grained spatial and logical constraints (\textit{e.g.}, lane topology, feasible maneuvers) within the reasoning procedure. This motivates approaches that directly couple multimodal reasoning with explicit structural semantics of the scene.

\subsection{Scene Understanding Benchmarks}
\paragraph{Low-level perception.}
\vspace{-0.4em}
Cityscapes~\cite{cordts2016cityscapes}, BDD100K~\cite{yu2018bdd100k}, and Mapillary Vistas~\cite{neuhold2017mapillary} provide large-scale 2D annotations for segmentation and detection. nuScenes~\cite{caesar2020nuscenes}, Waymo Open Dataset~\cite{sun2020scalability}, and Argoverse~\cite{chang2019argoverse,wilson2023argoverse} further extend to multimodal 3D boxes, tracking, and mapping. Although these datasets are foundational for geometry-centric and object-centric perception, they remain largely local and low-level, focusing on “what is where,” and they seldom encode mid-level semantics such as lane crossability, merge or diverge intent, or interpretable congestion as structured and interdependent tasks.
\vspace{-0.6em}
\paragraph{HD map construction and updating.}
\vspace{-0.6em}
Vectorized HD map methods~\cite{li2022hdmapnet,liu2023vectormapnet,liao2022maptr,liao2025maptrv2,xia2024dumapnet,xia2025ldmapnet-u} and multimodal-based road analysis approaches~\cite{xia2022duarus,xia2022dutraffic} predict road attributes such as instance changes, lane lines, road boundaries, and sometimes connectivity in a unified framework. Although effective, these approaches prioritize metrically accurate reconstruction from multi-sensor inputs, which is computationally expensive and labor-intensive to annotate. For many industrial scenarios (\textit{e.g.}, map freshness monitoring and change detection), lightweight camera-based semantic judgements are sufficient (\textit{e.g.}, lane-count change, new exit ramps), which are not the primary focus of existing HD map evaluations.

\vspace{-0.4em}
\paragraph{Multimodal and VLM-based driving benchmarks.}
\vspace{-0.4em}
Recent benchmarks such as NuScenes-QA~\cite{qian2024nuscenes}, NuInstruct~\cite{ding2024holistic}, DriveLM~\cite{sima2024drivelm}, VLADBench~\cite{li2025fine}, ROADWork~\cite{ghosh2025roadwork}, and DVBench~\cite{zeng2025vision} evaluate multimodal reasoning through VQA, instruction following, traffic-graph construction, or safety-critical scenarios. Although these datasets provide broad coverage, their labels are typically sparse or loosely coupled and seldom define per-frame mid-level attributes such as lane count or ego-lane index with explicit logical dependencies. As a consequence, they offer limited support for assessing whether models maintain a self-consistent, geometry-aware representation of local road topology, which is critical for reliable lane-level behavior reasoning.

\noindent\textbf{Summary.}
Prior work primarily focuses (i) low-level perception, (ii) HD map reconstruction, or (iii) high-level language-based reasoning. A clear gap remains in benchmarking lightweight, camera-based \emph{mid-level} road semantics with explicit structural dependencies, which are directly relevant to topology verification, map change detection, and the evaluation of geometry-aware multimodal reasoning (see Table~\ref{tab:benchmark_comparison}).

%% file: sec/3_dataset.tex
\section{RoadSceneBench}
\label{sec:approach}

The goal of RoadSceneBench is to evaluate and improve a model’s ability to understand mid-level road semantics, which include the structural and relational information that lies between perception and planning. Unlike existing datasets that emphasize dense labeling, RoadSceneBench centers on compact and interpretable tasks that encourage reasoning consistency rather than pixel-level precision.

\begin{figure*}[t]
    \centering
    \begin{subfigure}[t]{0.28\textwidth}
        \includegraphics[width=\linewidth]{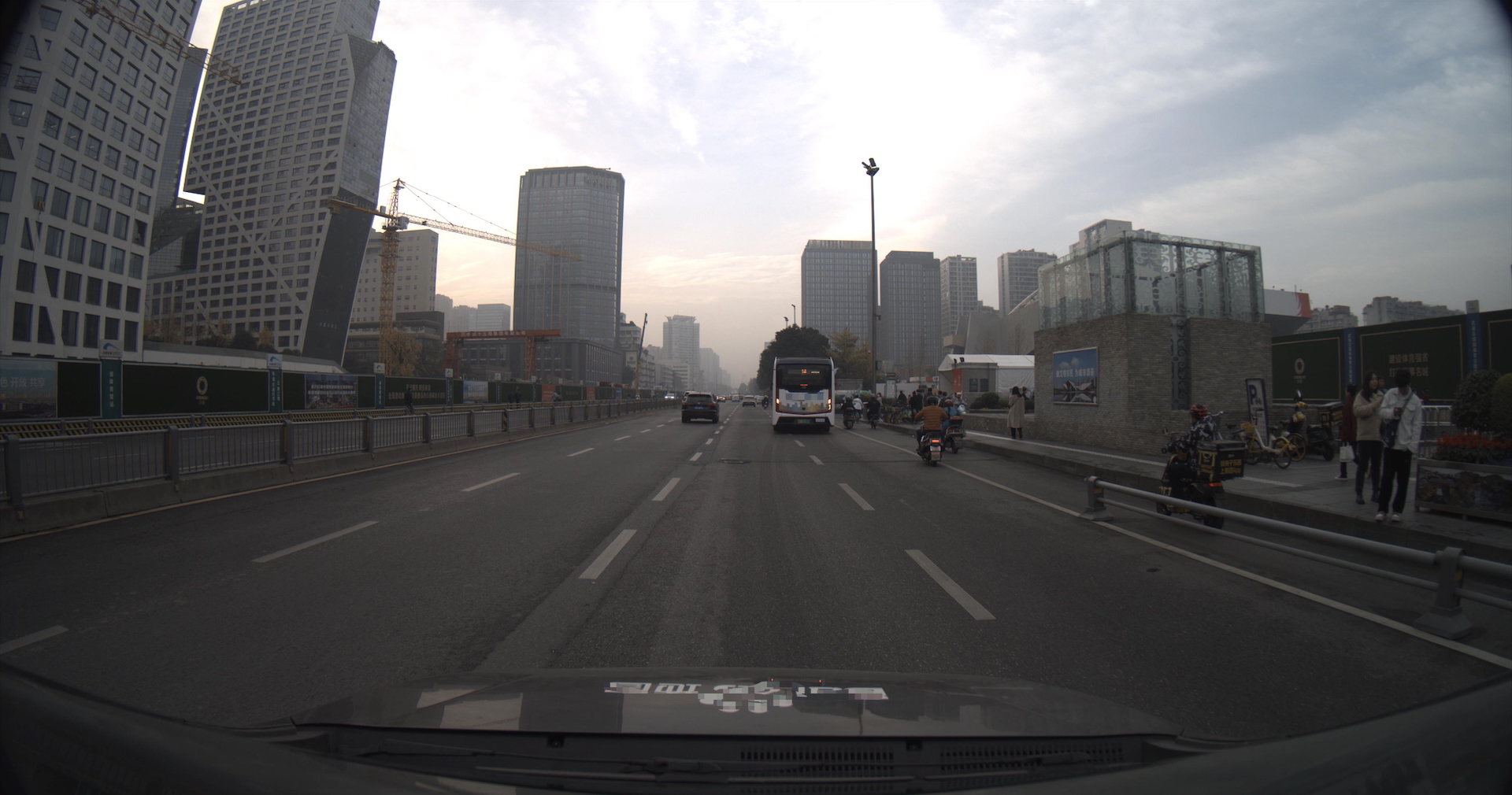}
        \caption{Urban}
    \end{subfigure}\hfill
    \begin{subfigure}[t]{0.28\textwidth}
        \includegraphics[width=\linewidth]{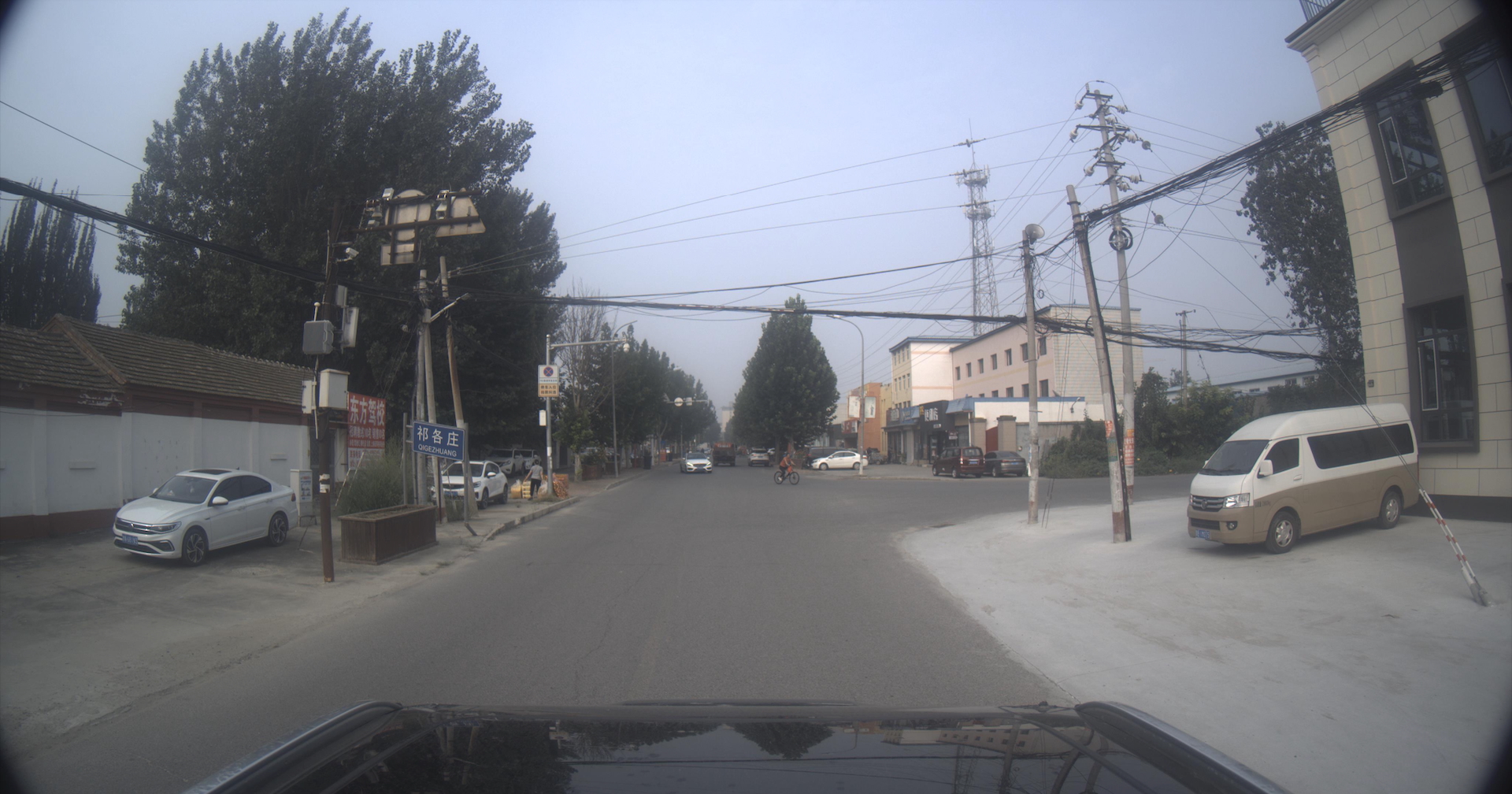}
        \caption{Suburban}
    \end{subfigure}\hfill
    \begin{subfigure}[t]{0.28\textwidth}
        \includegraphics[width=\linewidth]{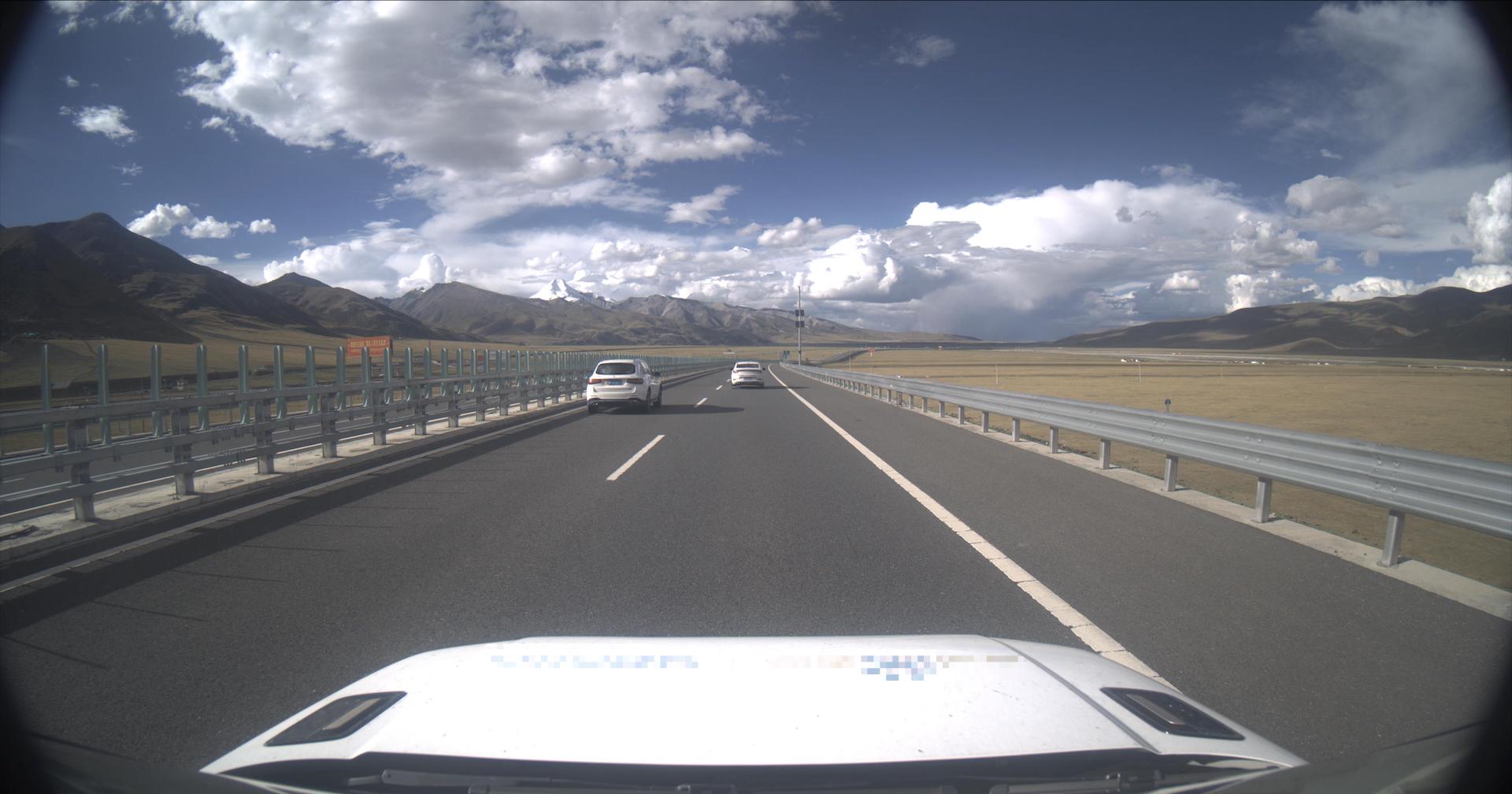}
        \caption{Highway}
    \end{subfigure}

    \begin{subfigure}[t]{0.28\textwidth}
        \includegraphics[width=\linewidth]{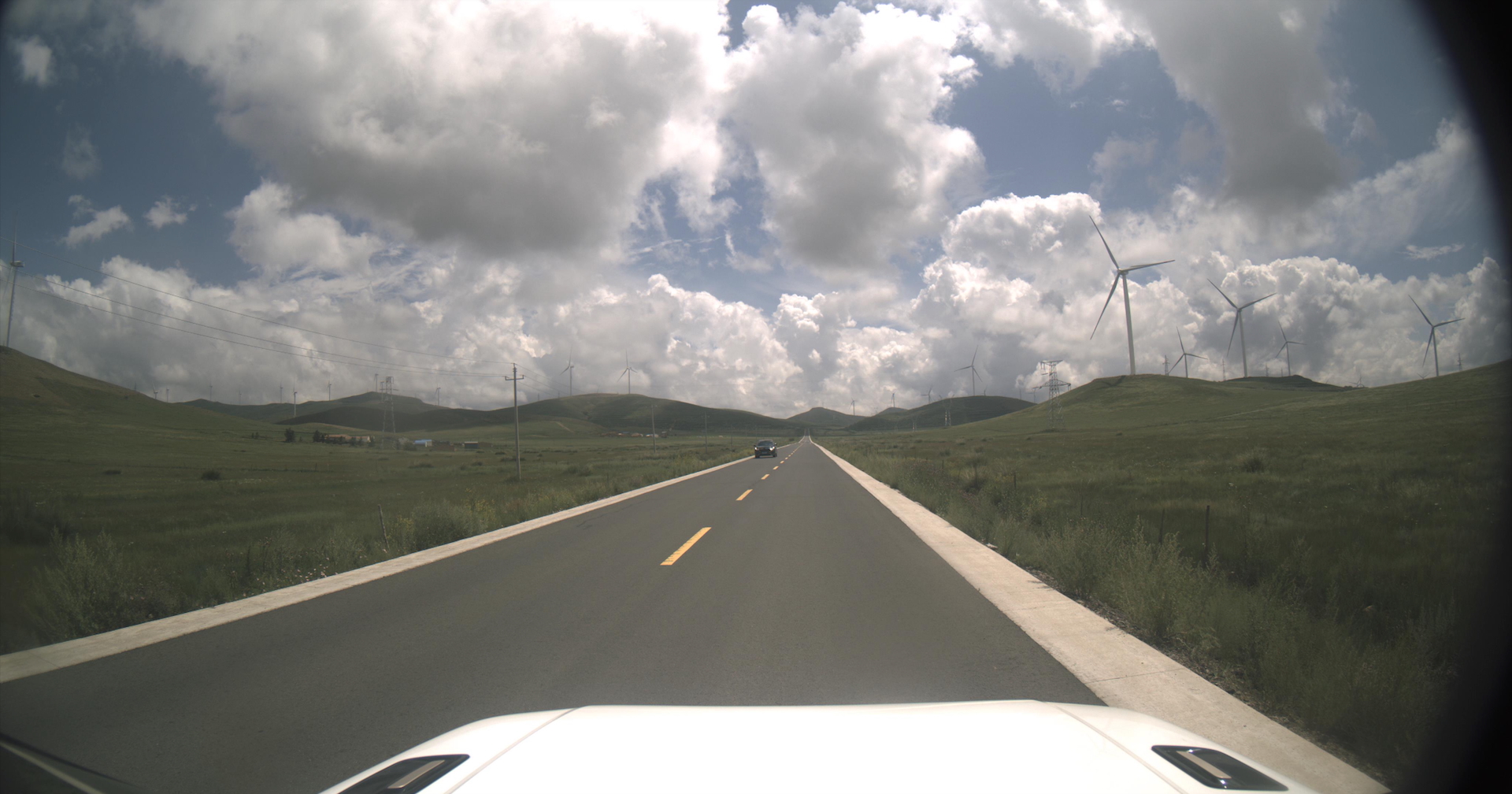}
        \caption{Free-flow}
    \end{subfigure}\hfill
    \begin{subfigure}[t]{0.28\textwidth}
        \includegraphics[width=\linewidth]{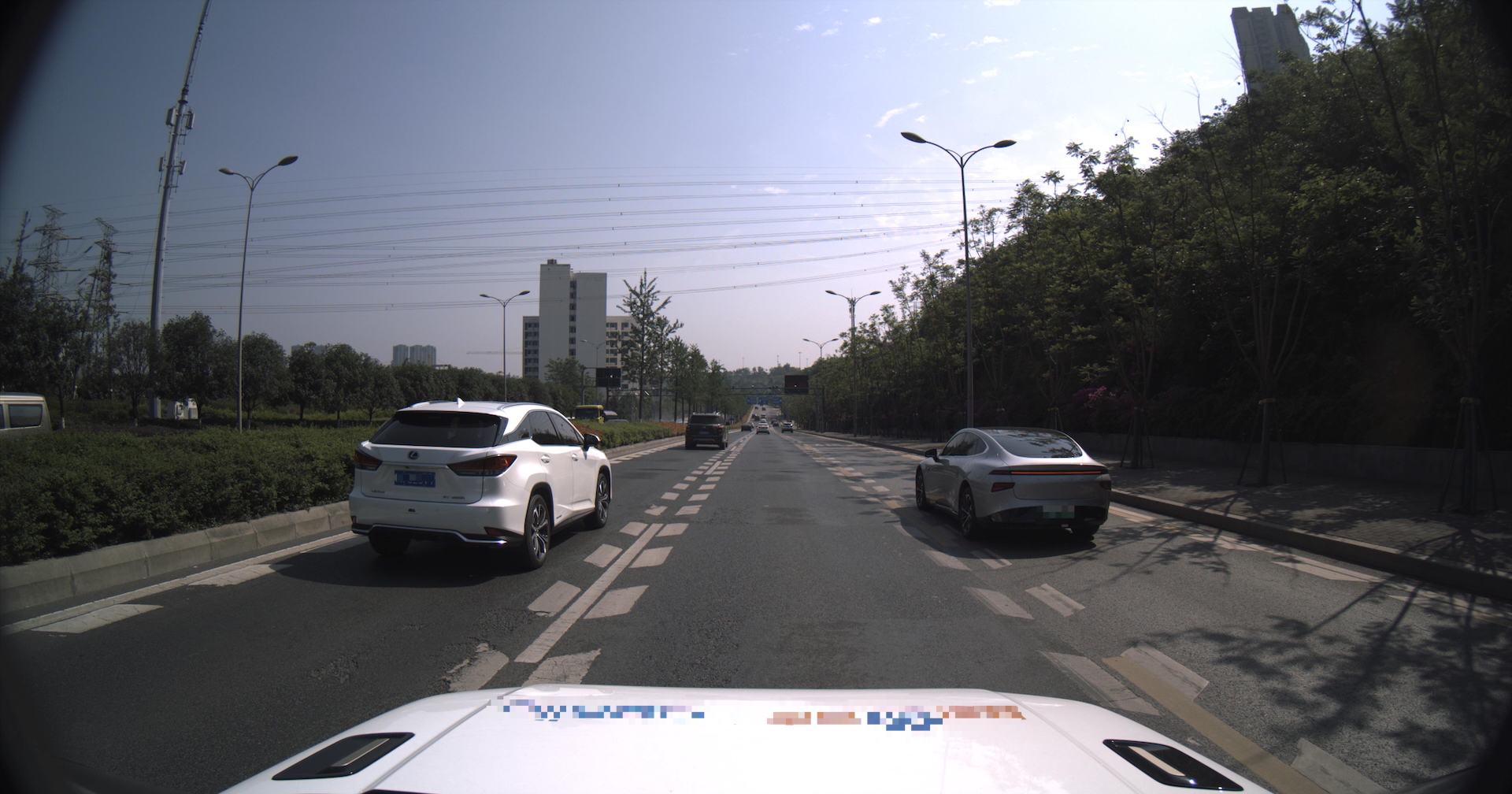}
        \caption{Moderate}
    \end{subfigure}\hfill
    \begin{subfigure}[t]{0.28\textwidth}
        \includegraphics[width=\linewidth]{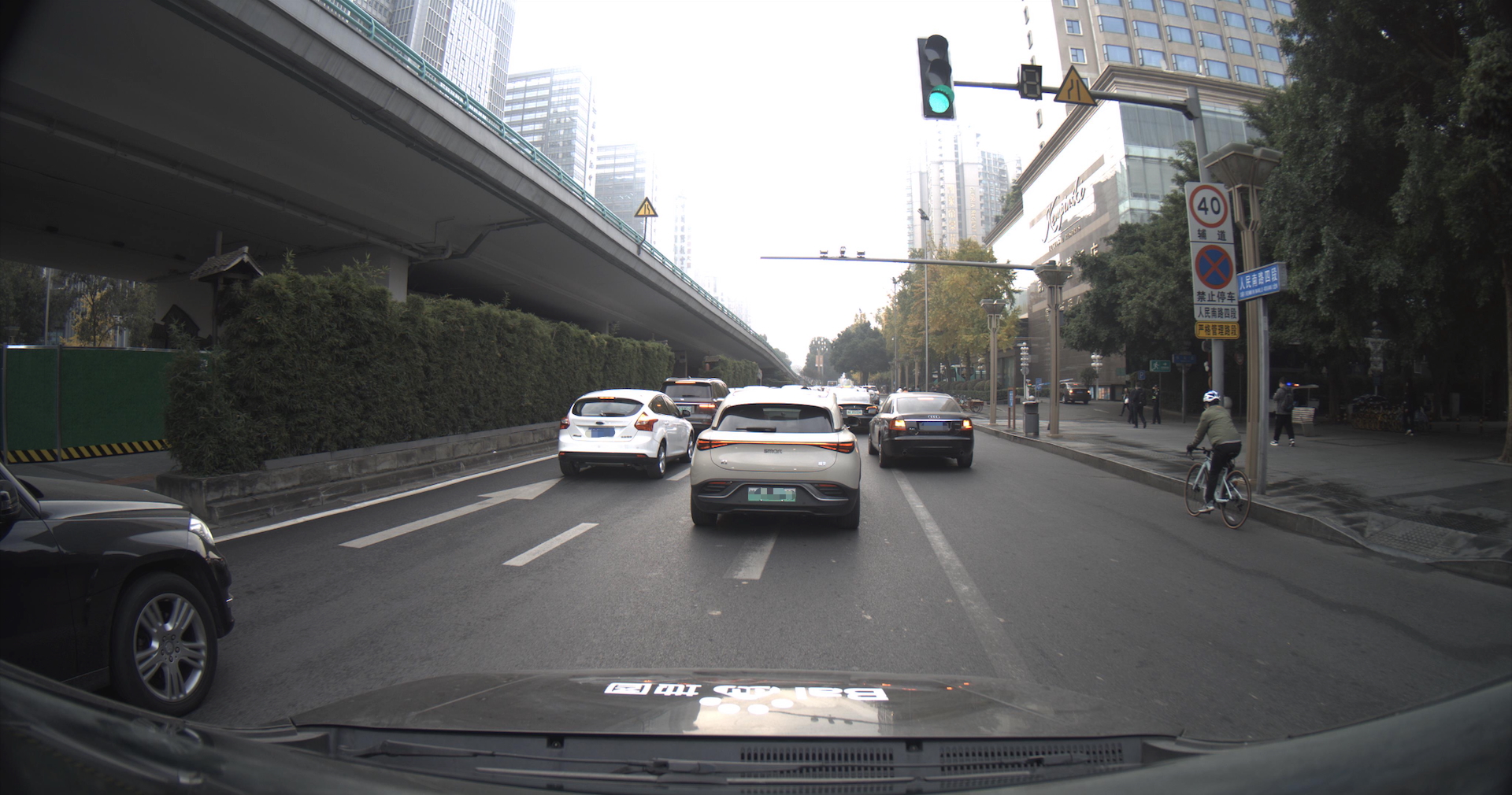}
        \caption{Congestion}
    \end{subfigure}

    \begin{subfigure}[t]{0.28\textwidth}
        \includegraphics[width=\linewidth]{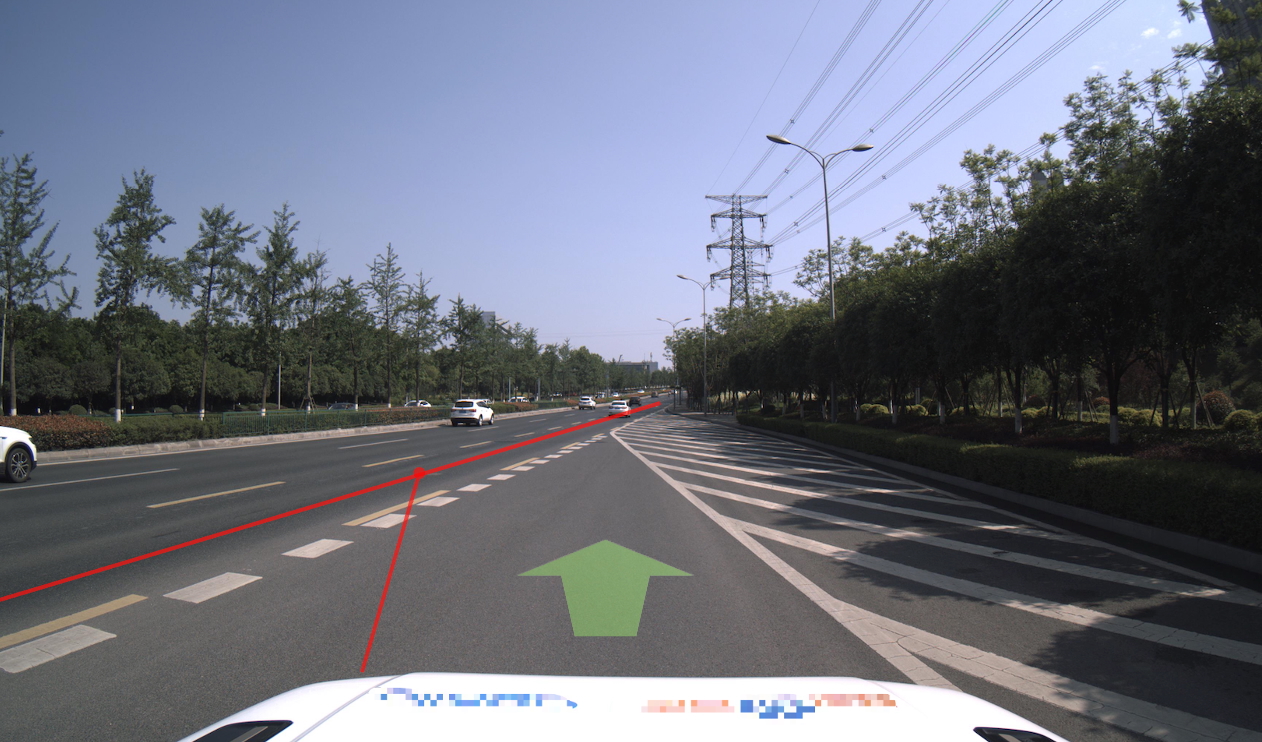}
        \caption{Entrance}
    \end{subfigure}\hfill
    \begin{subfigure}[t]{0.28\textwidth}
        \includegraphics[width=\linewidth]{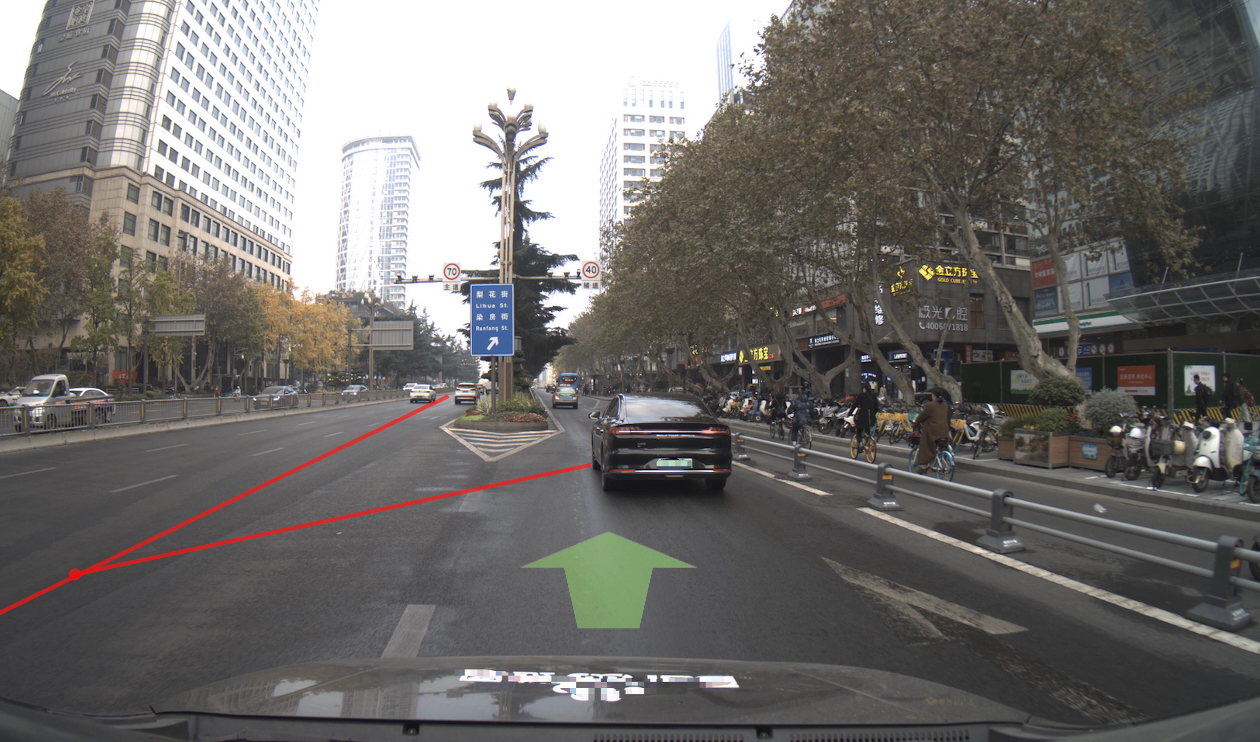}
        \caption{Exit}
    \end{subfigure}\hfill
    \begin{subfigure}[t]{0.28\textwidth}
        \includegraphics[width=\linewidth]{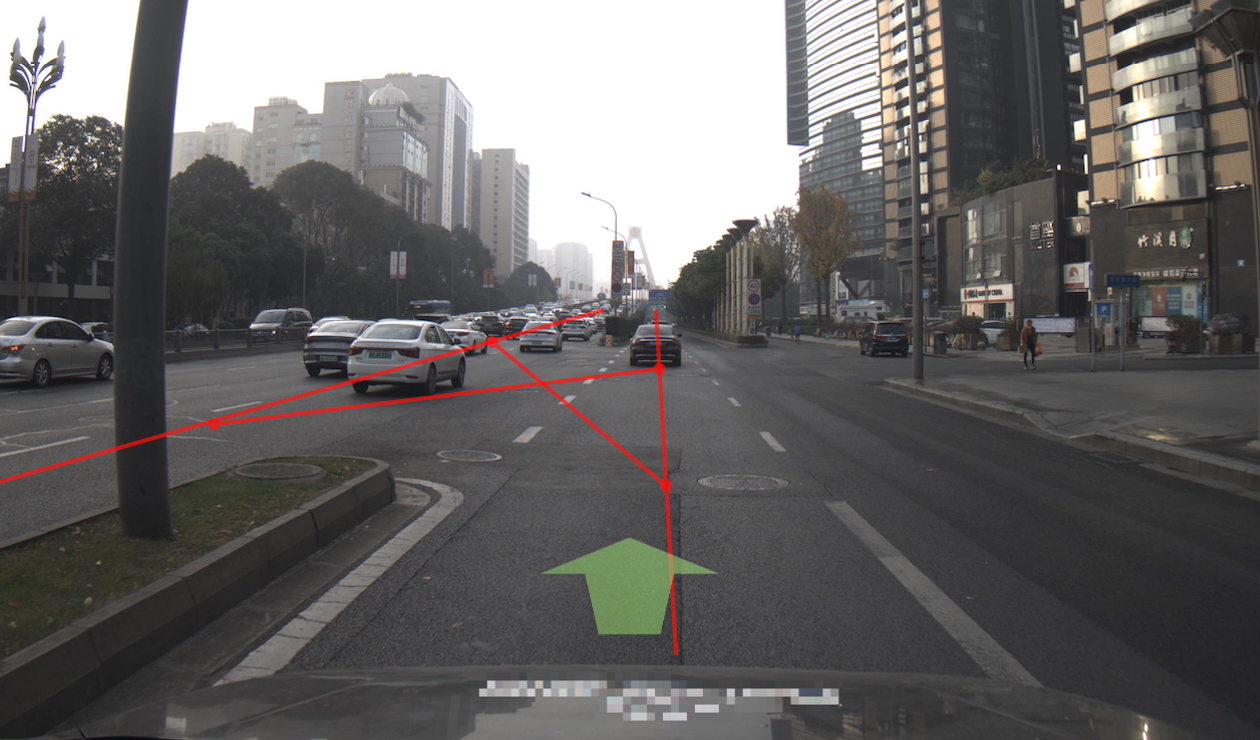}
        \caption{Junction}
    \end{subfigure}

    \begin{minipage}{0.16\textwidth}\hfill\end{minipage}
    \begin{subfigure}[t]{0.28\textwidth}
        \includegraphics[width=\linewidth]{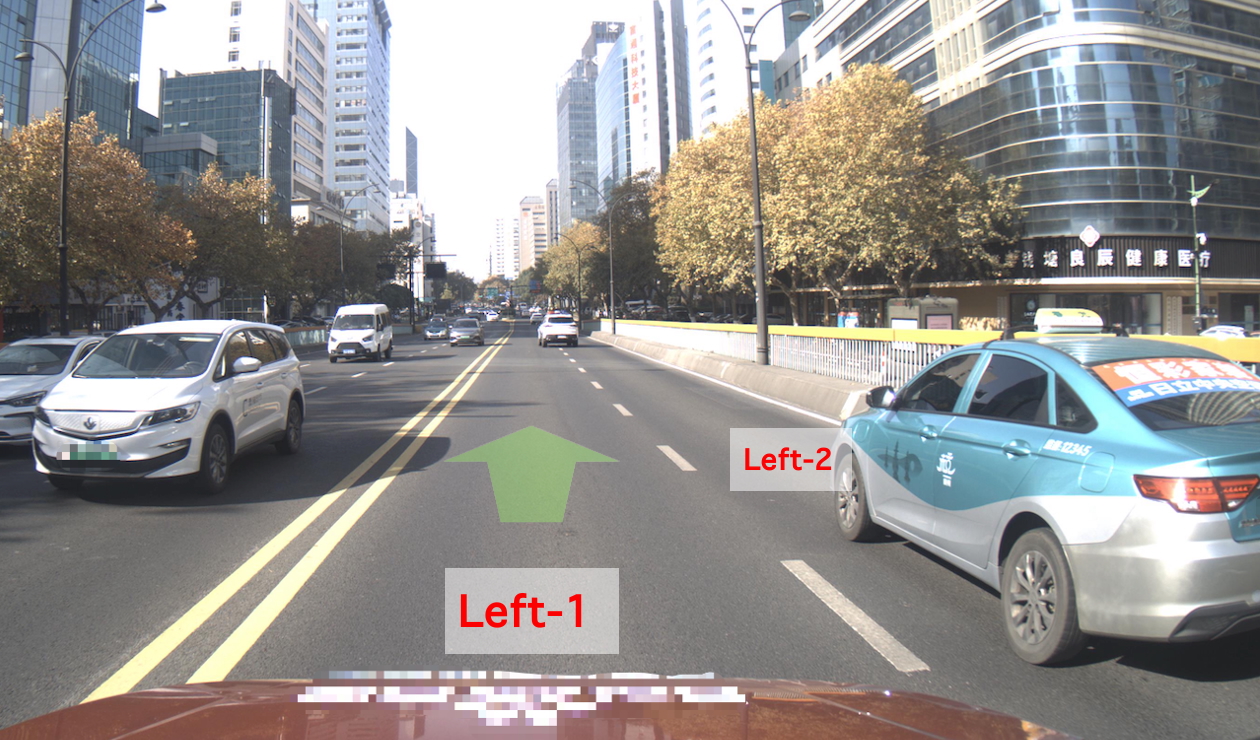}
        \caption{Lane Count \& Ego Lane Index}
    \end{subfigure}\hfill
    \begin{subfigure}[t]{0.28\textwidth}
        \includegraphics[width=\linewidth]{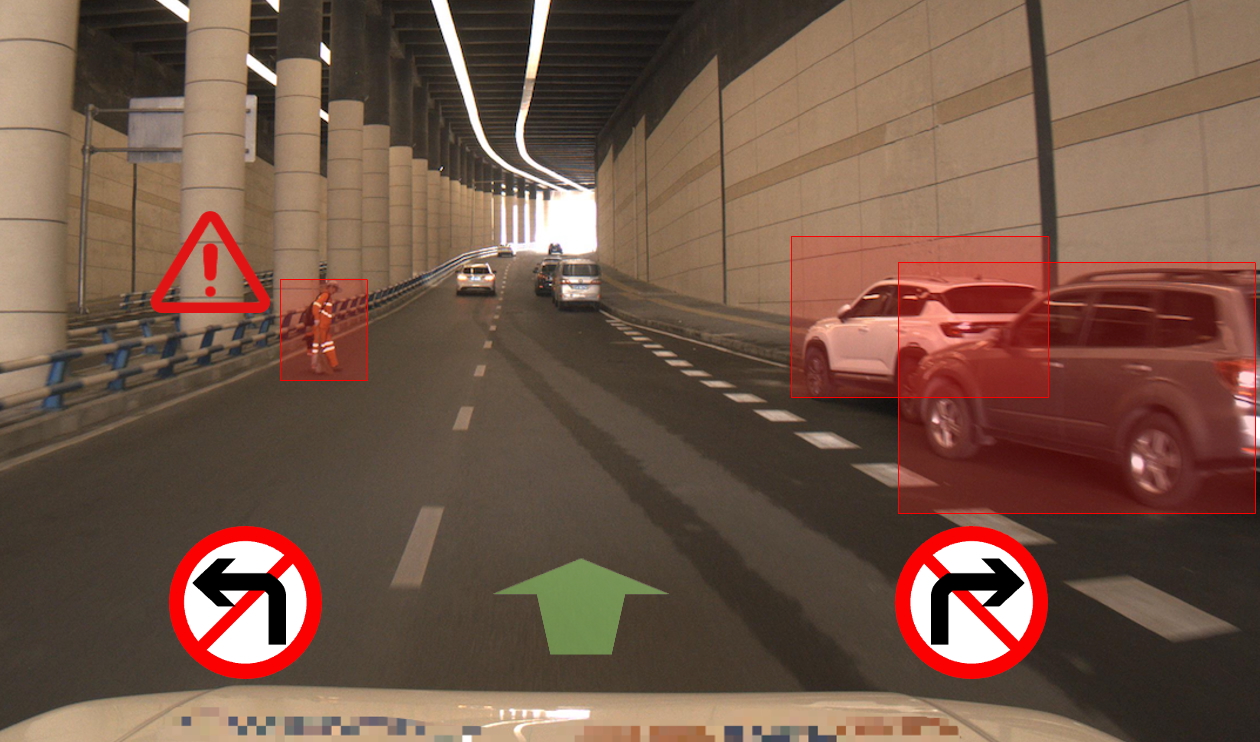}
        \caption{Lane-change Feasibility}
    \end{subfigure}
    \begin{minipage}{0.16\textwidth}\hfill\end{minipage}
    \vspace{-3mm}
    \caption{
        Visualization of representative annotation types in \textbf{RoadSceneBench}.
        All examples highlight the mid-level semantics connecting perception and structural reasoning.
    }
    \vspace{-4mm}
    \label{fig:benchmark_visualization}
\end{figure*}

\subsection{Tasks in Mid-level Scene Understanding}

RoadSceneBench defines relational reasoning tasks that reflect how road structures are perceived and interpreted from a driver’s perspective. Unlike conventional classification or detection benchmarks, these tasks emphasize the logical consistency among different road attributes. The tasks in RoadSceneBench are structurally interdependent rather than independent, forming a coherent reasoning framework. For instance, lane count constrains ego-lane index-if only three lanes exists, the vehicle cannot occupy a fourth. Similarly, entrance and exit cues influence connectivity reasoning, while traffic congestion often correlates with geometric complexity, such as in entrance ramps. This structured reasoning closely aligns with the mid-level representations used in industrial autonomous driving pipelines for high-definition (HD) map reconstruction and scene topology inference. See Figure~\ref{fig:benchmark_visualization} for visualizations.

\noindent\textbf{(i) Lane Count Estimation.} Predicting the number of visible lanes captures the global drivable structure of the scene. This task provides a structural prior for subsequent reasoning about connectivity or ego-lane index.

\noindent\textbf{(ii) Ego-lane Index.} Given the ego-vehicle’s view, the model predicts which lane it occupies. This establish a relative spatial frame that grounds all other relational tasks.

\noindent\textbf{(iii) Junction / Entrance / Exit Recognition.} Identifying whether the scene contains a junction or entrance/exit ramp. The task requires geometric and topological reasoning beyond pixel-level cues and aligns closely with lane-level topology reconstruction.

\noindent\textbf{(iv) Lane-change Feasibility Reasoning.} Assessing whether the ego vehicle can legally and physically transition into an adjacent lane. This task requires reasoning over lane-boundary type, geometric continuity, and local traffic context, and is closely related to lane-level topology inference and drivable-region connectivity in map construction.

\noindent\textbf{(v) Traffic Condition Understanding.} Predicting whether the scene exhibits free flow, moderate, or congestion. Unlike typical traffic flow estimation tasks relying on sensor data, this relies purely on visual context and spatial priors, which provides contextual cues for maneuver prediction.

\noindent\textbf{(vi) Road Scene Classification.} Classifying the driving environment into urban, suburban, or highway scenes based on global roadway layout and surrounding context.

\subsection{Dataset Construction}
\noindent\textbf{Data Collection.} To build a benchmark that enables fine-grained and comprehensive evaluation of mid-level street-scene reasoning under diverse and challenging conditions, we collected raw street-view imagery using a fleet of vehicles equipped with professional-grade sensors and cameras. Due to policy constraints, the data collection was restricted to regions within China. To ensure diversity in both geographic coverage and scene structure, we selected $20$ representative cities, including Beijing, Shanghai, and others, capturing a wide spectrum of urban scales, spatial distributions, and road network configurations.

\noindent\textbf{Data Preparation.} We initially gathered over $100,000$ images, followed by a two-stage filtering pipeline to obtain high-quality and representative samples. In the first stage, a suite of in-house models for classification and segmentation was used to automatically remove low-quality samples, such as overexposed images, invalid scenes, and duplicates. In the second stage, a team of $20$ professionally trained annotators manually reviewed the remaining images to ensure semantic diversity and annotation reliability. Ultimately, we curated $2,341$ short clips, each containing $5$ consecutive frames, capturing temporal continuity while maintaining a manageable annotation workload.

\noindent\textbf{Q\&A Annotation.} To balance annotation cost and quality, we adopted a semi-automated labeling strategy. We first employed a set of pre-trained classification and segmentation models from our previous work to generate pseudo-labels, which were subsequently reviewed and corrected by expert annotators. Annotators were instructed to enforce logical consistency across tasks and temporal coherence within frame sequences. The annotation team followed a question–answer protocol aligned with the six core tasks defined in RoadSceneBench. For example, if the ego vehicle is positioned in the leftmost lane, a left lane-change action should not be permitted. Given the highly structured nature of mid-level road-scene understanding, we further designed a template-based question framework that integrates automated pre-labels with expert reasoning, ensuring semantic accuracy and structural coherence across tasks.

\begin{figure}
\includegraphics[width=1.0\linewidth]{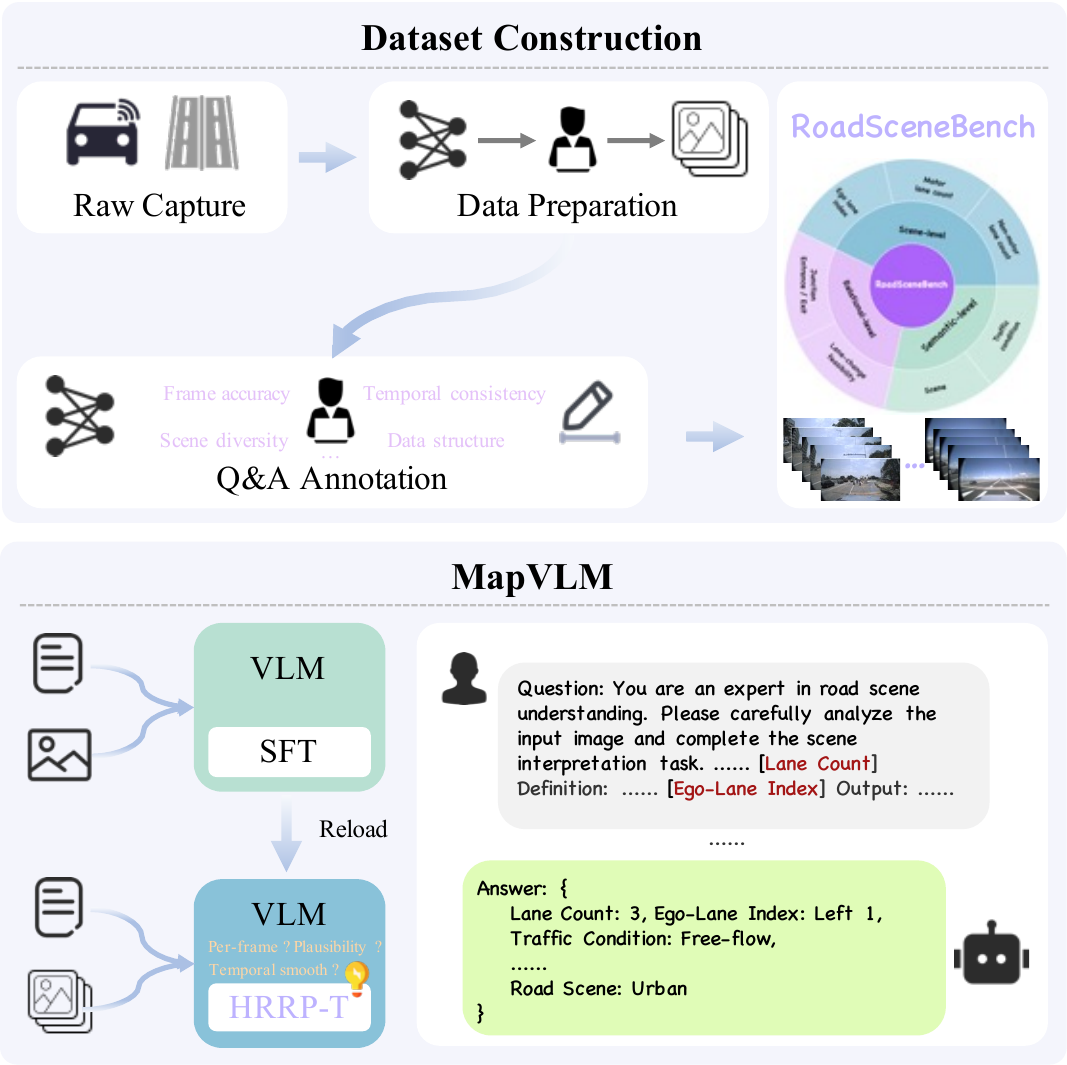}
\caption{Framework of dataset construction and MapVLM.}
\vspace{-5mm}
\label{fig:framework}
\end{figure}

\subsection{Statistical Analysis}
Our RoadSceneBench dataset is organized into six major task categories. It contains $11,705$ high-resolution images ($4096 \times 2160$ pixels) forming $2,341$ clips, each composed of five temporally consecutive frames sampled at one frame per second. In total, the dataset provides over $163k$ annotated labels collected from $20$ cities, including Beijing, Shanghai, and others, covering diverse urban environments such as city centers, suburban areas, and highways. 

%% file: sec/4_method.tex
\section{MapVLM}
\label{sec:approach}

To enable fine-grained understanding of mid-level road semantics, we introduce MapVLM (see Figure~\ref{fig:framework}), a vision–language model training paradigm tailored for RoadSceneBench. MapVLM integrates Supervised Fine-Tuning (SFT) with Hierarchical Relational Reward Propagation with Temporal Consistency (HRRP-T), a reinforcement learning framework that progressively aligns visual representations, linguistic descriptions, and structural priors, thereby bridging the gap between static perception and dynamic reasoning.

\subsection{Supervised Fine-Tuning}

In the first stage, we adopt Qwen2.5-VL-7B~\cite{qwen2.5-vl} as the base model and apply LoRA-based~\cite{hu2022lora} fine-tuning to establish core reasoning capabilities for mid-level road-scene understanding. Each training sample consists of a single image paired with a structured query derived from the benchmark tasks. The resulting model is able to produce both direct answers and structured scene descriptions, including lane count, ego-lane index, junction/entrance/exit presence, lane-change feasibility, traffic condition, and road scene type. This stage provides initial alignment between visual semantics and language grounding; however, it lacks inter-task and temporal consistency. To overcome this limitation, we introduce HRRP-T, a reinforcement learning paradigm designed to enforce multi-level relational coherence across spatial, temporal, and semantic dimensions.

\subsection{Hierarchical Relational Reward Propagation}
In the second stage, we extend the Hierarchical Relational Reward Propagation framework into a temporally aware variant, HRRP-T, to enhance the model’s robustness and reasoning continuity in dynamic road environments. Unlike frame-independent evaluation, HRRP-T explicitly models short temporal sequences through two complementary reward components: a frame-level reward that evaluates per-frame accuracy across scene-level, relational-level, and semantic-level layers, and a temporal-level reward that enforces inter-frame consistency by promoting plausible temporal transitions and penalizing abrupt or unrealistic changes. By coupling these signals, HRRP-T encourages both spatial precision and temporal coherence, which are two essential properties for VLMs operating in real-world road environments.

\subsubsection{Frame-Level Hierarchical Reward Design}

At the frame level, HRRP-T adopts a three-layer hierarchical structure consisting of the scene-level, relational-level, and semantic-level components. Scene-level Layer evaluates low-level spatial topology, such as lane count and ego-lane index, by comparing model predictions with frame-wise ground-truths. Relational-level Layer captures mid-level reasoning over spatial and functional relationships, including the recognition of junctions and entrance/exit structures, as well as the inference of lane-change feasibility based on static lane-boundary constraints (\textit{e.g.}, solid lines) and dynamic obstacles (\textit{e.g.}, vehicles, pedestrians). Semantic-level Layer measures high-level scene understanding, such as road-scene type and traffic condition. Each frame produces a hierarchical reward vector that jointly quantifies spatial and semantic correctness across these three layers:
\begin{equation}
\label{equ1}
\mathcal{R}_{frame}^{t}=\alpha R_{sce}^{t}+\beta \mathcal{R}_{rel}^{t}+\gamma \mathcal{R}_{sem}^{t},
\end{equation}
where $t$ denotes the $t^{th}$ frame within a clip.

\subsubsection{Temporal-Level Hierarchical Reward Design}

Real-world road environments are inherently non-stationary: vehicles move, lanes merge or diverge, and occlusions naturally arise. To account for such variability, the temporal-level reward in HRRP-T does not assume strict frame-to-frame continuity; instead, it evaluates whether the predicted temporal evolution is plausible within a short temporal horizon. To this end, we introduce a two-level temporal consistency reward, denoted as $\mathcal{R}_{temporal}$, which decomposes temporal dynamics into two complementary components—local smoothness and semantic plausibility—jointly capturing short-term stability and high-level temporal coherence:
\begin{equation}
\label{equ2}
\mathcal{R}_{temporal}=\lambda \mathcal R_{smooth}+(1-\lambda) \mathcal R_{plausible},
\end{equation}
where $\lambda \in [0,1]$ balances the contribution between smoothness and plausibility.

The smoothness reward, $\mathcal R_{smooth}$, encourages gradual and consistent variation across adjacent frames, penalizing abrupt or oscillatory predictions. Formally, it is defined as:
\begin{equation}
\label{equ3}
\mathcal{R}_{smooth}=1-\frac{1}{T-1}\sum_{t=1}^{T-1}|y_{t}-y_{t-1}|,
\end{equation}
where $y_{t}$ denotes the prediction at frame $t$, and $T$ is the total number of frames. This term mainly regularizes continuous or ordinal variables, such as lane count, by encouraging their temporal evolution to be smooth rather than exhibiting frame-by-frame discontinuities. A high smoothness reward is assigned when consecutive predictions vary gradually (\textit{e.g.}, $3 \rightarrow 2 \rightarrow 2$), whereas unstable or erratic fluctuations (\textit{e.g.}, $3 \rightarrow 1 \rightarrow 3$) yield a lower reward.

While smoothness encourages temporal continuity, it does not by itself ensure that transitions are semantically valid. To complement this, the plausibility reward $\mathcal{R}_{plausible}$ evaluates whether temporal changes adhere to physically and semantically reasonable transitions:
\begin{equation}
\label{equ4}
\mathcal R_{\text{plausible}} = \frac{1}{T - 1} \sum_{t=1}^{T-1} \mathbb{I}\big(\text{V}(y_t, y_{t+1})\big),
\end{equation}
where $\mathbb{I}(\cdot)$ is the indicator function:
  \[
  \mathbb{I}(\text{condition}) =
  \begin{cases}
  1, & \text{if the condition holds true},\\
  0, & \text{otherwise};
  \end{cases}
  \],
and $\text{V}(y_t, y_{t+1})$ is a logical function that determines whether the transition from $y_t$ to $y_{t+1}$ adheres to semantically valid, as specified by domain priors or empirical statistics. For instance, in the case of lane-change feasibility, transitions from “changeable” to “non-changeable” (\textit{e.g.}, due to encountering a solid line) are acceptable, whereas rapid oscillations between the two states are discouraged. This component effectively introduces a lightweight finite-state machine~\cite{hopcroft2001introduction}–like constraint into the temporal domain, ensuring that frame-wise predictions remain not only smooth but also semantically and physically coherent. Finally, HRRP-T integrates both frame-level and temporal-level rewards and is trained using the GRPO strategy~\cite{guo2025deepseek} :
\begin{equation}
\label{equ5}
\mathcal R_{\text{HRRP-T}} = \lambda_{frame}\frac{1}{T} \sum_{t=1}^{T} \mathcal R_{frame}^{t} +  \lambda_{temporal} \mathcal R_{temporal},
\end{equation}
where $\lambda_{frame}$ and $\lambda_{temporal}$ are the hyper-parameters to balance the trade-off between two terms.

\begin{table*}
\centering
\caption{Quantitative comparison of baseline VLMs (closed- and open-source) against our MapVLM on the RoadSceneBench.}
\vspace{-2mm}
\label{tab:main_results}
\resizebox{2.1\columnwidth}{!}{
\begin{tabular}{@{}lcccccccccccc|cc@{}}
\toprule
\textbf{Metric}             & \multicolumn{2}{c}{\textbf{Lane Count}} & \multicolumn{2}{c}{\textbf{Ego-lane Index}} & \multicolumn{2}{c}{\textbf{\begin{tabular}[c]{@{}c@{}}Lane Change\\ Feasibility\end{tabular}}} & \multicolumn{2}{c}{\textbf{Traffic Condition}} & \multicolumn{2}{c}{\textbf{Road Scene}} & \multicolumn{2}{c|}{\textbf{Road Topology}} & \multicolumn{2}{c}{\textbf{Overall}} \\ \midrule
                            & P(\%)              & R(\%)              & P(\%)                & R(\%)                & P(\%)                                          & R(\%)                                         & P(\%)                  & R(\%)                 & P(\%)              & R(\%)              & P(\%)                & R(\%)                & P(\%)        & R(\%)        \\ \cmidrule(r){1-1}
\textit{Closed-Source VLMs} &                    &                    &                      &                      &                                                &                                               &                        &                       &                    &                    &                      &                      &              &              \\
GPT-4o                      & 51.02              & 32.39              & 23.58                & 24.50                & 42.21                                          & 35.57                                         & 78.76                  & 60.39                 & 67.64              & 64.23              & 47.50                & 35.42                & 51.79        & 42.08        \\
Gemini-2.5-pro              & 52.78              & 43.06              & 72.71                & 46.52                & 59.25                                          & 52.98                                         & 73.69                  & 59.26                 & 65.91              & 64.20              & 39.33                & 50.20                & 60.61        & 52.70        \\
Claude-3.7V Sonnet          & 28.64              & 28.34              & 27.53                & 25.24                & 41.11                                          & 44.89                                         & 80.17                  & 51.89                 & 61.58              & 59.25              & 44.75                & 38.68                & 47.30        & 41.38        \\ \cmidrule(r){1-1}
\textit{Open-Source VLMs}   &                    &                    &                      &                      &                                                &                                               &                        &                       &                    &                    &                      &                      &              &              \\
ERNIE-4.5-VL-28B-A3B        & 32.25              & 28.39              & 26.46                & 12.72                & 27.09                                          & 21.41                                         & 79.38                  & 47.70                 & 69.98              & 60.88              & 65.48                & 32.49                & 50.11        & 33.93        \\
DeepSeek-VL2                & 9.80               & 10.44              & 18.36                & 12.10                & 7.17                                           & 16.14                                         & 19.05                  & 16.31                 & 28.50              & 19.26              & 16.88                & 13.18                & 16.63        & 14.57        \\
LLaVA-Onevision / Next-7B   & 37.75              & 45.84              & 30.07                & 27.75                & 38.58                                          & 28.97                                         & 59.56                  & 61.14                 & 42.80              & 41.63              & 23.11                & 25.92                & 38.65        & 38.54        \\
InternVL3-1B                & 22.06              & 2.88               & 11.98                & 25.00                & 5.93                                           & 19.25                                         & 41.76                  & 38.02                 & 52.67              & 34.51              & 22.28                & 19.46                & 26.11        & 23.19        \\
InternVL3-8B                & 29.80              & 28.59              & 28.86                & 30.59                & 2.90                                           & 20.00                                         & 54.68                  & 37.22                 & 73.07              & 69.32              & 46.15                & 27.34                & 39.24        & 35.51        \\
InternVL3-78B               & 53.42              & 36.80              & 28.97                & 25.40                & 50.86                                          & 47.66                                         & 81.05                  & 53.71                 & 69.02              & 69.92              & 49.83                & 38.10                & 55.53        & 45.27        \\
Qwen2.5-VL-3B               & 29.65              & 22.89              & 9.73                 & 19.58                & 13.66                                          & 26.35                                         & 70.54                  & 36.09                 & 71.78              & 64.79              & 48.32                & 33.46                & 40.61        & 33.86        \\
Qwen2.5-VL-7B               & 12.81              & 20.66              & 27.16                & 18.56                & 34.34                                          & 26.09                                         & 71.92                  & 40.35                 & 69.79              & 68.51              & 39.98                & 26.04                & 42.67        & 33.37        \\
Qwen2.5-VL-32B              & 33.20              & 29.79              & 30.56                & 25.18                & 47.73                                          & 31.47                                         & 64.94                  & 46.74                 & 68.07              & 63.02              & 57.92                & 28.77                & 50.40        & 37.50        \\
Qwen2.5-VL-72B              & 30.23              & 24.68              & 37.03                & 38.07                & 39.44                                          & 31.70                                         & 72.68                  & 46.93                 & 67.85              & 67.40              & 56.16                & 32.29                & 50.57        & 40.18        \\
Qwen3-VL-2B                 & 36.01              & 33.59              & 30.74                & 31.14                & 4.31                                           & 25.07                                         & 55.78                  & 49.14                 & 52.59              & 50.19              & 37.46                & 41.97                & 36.15        & 38.52        \\
Qwen3-VL-8B                 & 54.99              & 34.93              & 29.78                & 31.45                & 47.16                                          & 40.90                                         & 80.46                  & 54.78                 & 72.79              & 70.15              & 58.84                & 30.70                & 57.34        & 43.82        \\
Qwen3-VL-32B                & 50.30              & 34.99              & 36.28                & 33.73                & 40.97                                          & 43.93                                         & 78.79                  & 50.24                 & 65.53              & 64.16              & 49.38                & 35.10                & 53.54        & 43.69        \\ \cmidrule(r){1-1}
\textbf{MapVLM (SFT)}      & 66.02                & 61.55                & 69.34                  & 50.37                  & 87.63                                            & 88.27                                           & 75.68           & 74.87          & 81.62                & 82.65                & 52.56                  & 45.81                  & 72.14          & 67.25          \\
\textbf{MapVLM (SFT+HRRP-T)}      & 63.37                & 65.89                & 75.44                  & 84.67                  & 83.83                                            & 84.69                                           & 82.35           & 72.70          & 82.80                & 82.36                & 66.88                  & 42.73                  & \textbf{75.78}          & \textbf{72.17}          \\ \bottomrule
\end{tabular}
}
\end{table*}
\vspace{-2mm}

\section{Experiments}

\subsection{Experimental Settings}

For our comparative study, we employe three powerful commercial Closed-Source VLMs as baseline models: GPT-4o~\cite{gpt-4o}, Gemini-2.5-Pro~\cite{gemini-2.5-pro}, and Claude-3.7-Sonnet~\cite{claude-3.7-sonnet}. These models represent the state-of-the-art capability among current proprietary VLMs. Furthermore, our investigation include a comprehensive set of 12 Open-Source VLMs spanning five major families: ERNIE~\cite{ernie-4.5-vl}, DeepSeek~\cite{deepseek-vl2}, LLaVA~\cite{llava-onevision}, InternVL~\cite{internvl3}, and Qwen series~\cite{qwen2.5-vl,qwen3-vl}. For the quantitative evaluation of aforementioned six tasks, we employ \textbf{Precision} (P) and \textbf{Recall} (R) as the primary metrics. For the closed-source VLMs, we conducted zero-shot evaluations by querying their official APIs. For the open-source VLMs, we utilized the \textit{swift}~\cite{swift} framework to facilitate both fine-tuning and zero-shot testing. All experiments involving open-source models were executed on a computing cluster provisioned with A800 GPUs. Across all evaluation inferences, we employed a deterministic decoding strategy by setting the \textit{temperature} to 0.0 and \textit{top\_p} to 1.0.

\subsection{Main Results}
The comprehensive performance of all baseline VLMs, alongside our proposed MapVLM, on the various tasks within RoadSceneBench is detailed in Table~\ref{tab:main_results}. Our analysis of these results yields two primary insights:

\noindent\textbf{Challenges of the RoadSceneBench.}
First, the results in Table~\ref{tab:main_results} underscore the challenging nature of our benchmark. We observe that two tasks in particular, `Ego-lane Index' and `Lane Change Feasibility', are exceptionally difficult. A significant number of both closed-source and open-source VLMs achieve relatively low Precision (P) and Recall (R) on these metrics. This finding is critical, as these two tasks are arguably the most pertinent to real-world autonomous driving perception and decision-making. Furthermore, we note a high degree of performance variance for individual models across different tasks. For example, {Qwen2.5-VL-3B} shows competent performance on {`Road Scene Categorization'} (P: 71.78\%, R: 64.79\%) but struggles significantly with {`Ego-lane Index'} (P: 9.73\%, R: 19.58\%), highlighting the specialized and uneven capabilities of current models.

\noindent\textbf{Model Comparison.}
When comparing model families based on the `Overall' metrics, the {closed-source VLMs generally outperform their open-source counterparts}.
Among the closed-source VLMs, {Gemini-2.5-Pro} emerges as the top-performing baseline, achieving the comprehensive scores with an Overall Precision of 60.61\% and Recall of 52.70\%.
Among the open-source VLMs, the performance is more varied. {Qwen3-VL-8B} achieves the {highest Overall Precision} (57.34\%), while {InternVL3-78B} secures the {highest Overall Recall} (45.27\%). 
This demonstrates a clear trade-off between precision and recall within the open-source landscape. 

{In stark contrast, our {MapVLM} demonstrates significantly superior performance, establishing a new SOTA on the {RoadSceneBench}. 
As shown in Table~\ref{tab:main_results}, MapVLM achieves an Overall Precision of 75.78\% and Recall of 72.17\%, decisively surpassing the best-performing baseline ({Gemini 2.5 Pro}) by a substantial margin.}
{Furthermore, this superiority is consistent across all six evaluation tasks. 
MapVLM not only achieves the top P and R scores in almost tasks, but it particularly excels in the most challenging tasks identified earlier: `Ego-lane Index' and `Lane Change Feasibility'. 
This comprehensive dominance underscores the effectiveness of our model training strategies and its advanced reasoning capabilities for complex road topology and dynamic scenarios.}

\begin{figure*}[t] 
	\centering
	\center{\includegraphics[width=16.5cm]{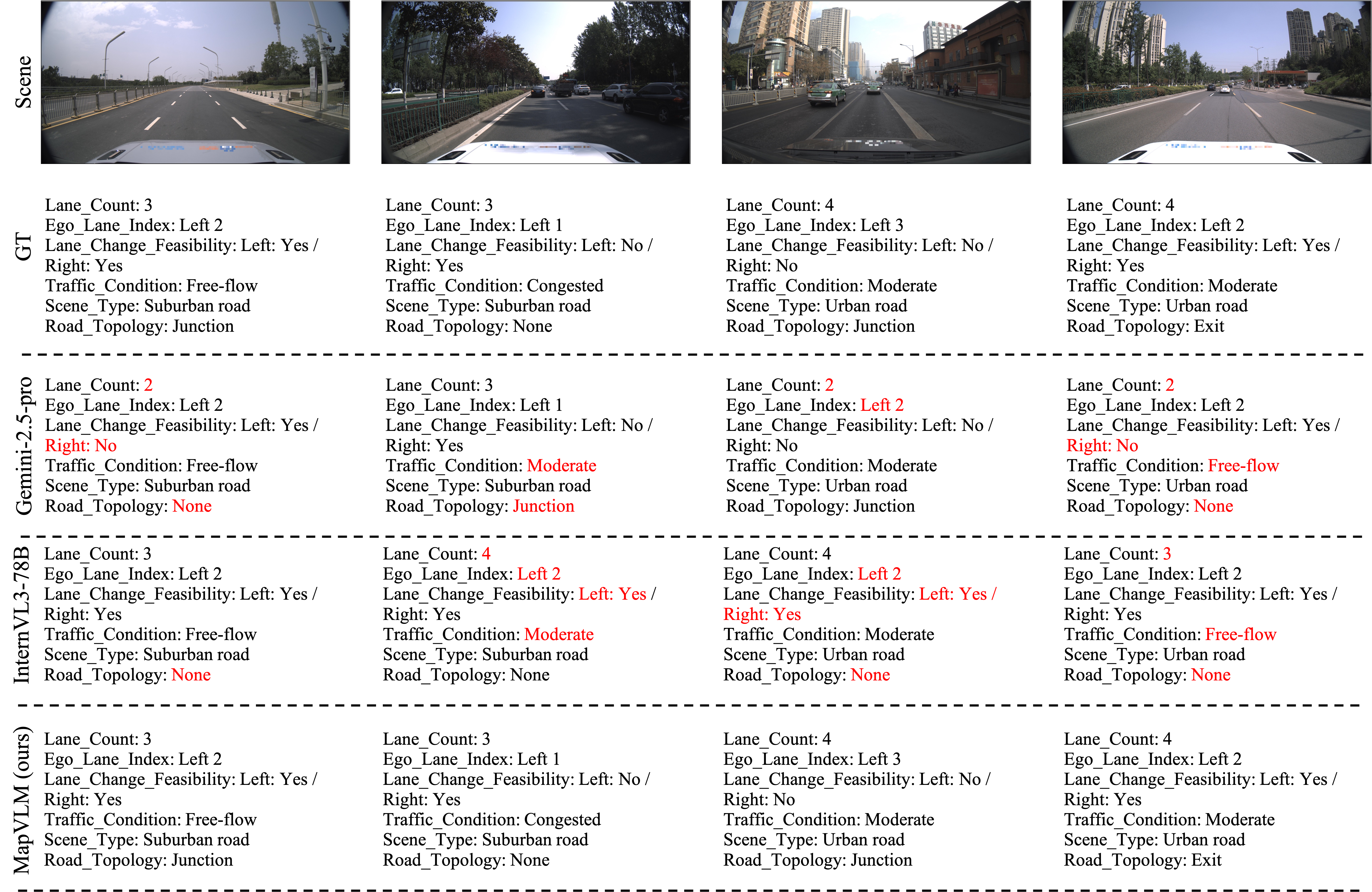}}
    \vspace{-2mm}
	\caption{Visualization of representative cases from {RoadSceneBench} that highlight the complexities of real-world driving scenarios. {Red text} indicates prediction errors, underscoring both the benchmark's difficulty and the superior reasoning of our model.}\label{visualize_0}
    \vspace{-2mm}
\end{figure*}

\subsection{Visualization}
To provide an intuitive understanding of our benchmark's complexity and the comparative performance of various models, we present several representative examples in Figure~\ref{visualize_0}.

\begin{figure*}[t]
    \vspace*{-3pt} 
    \centering

    \begin{minipage}{0.19\textwidth}
        \centering
        \includegraphics[width=\linewidth]{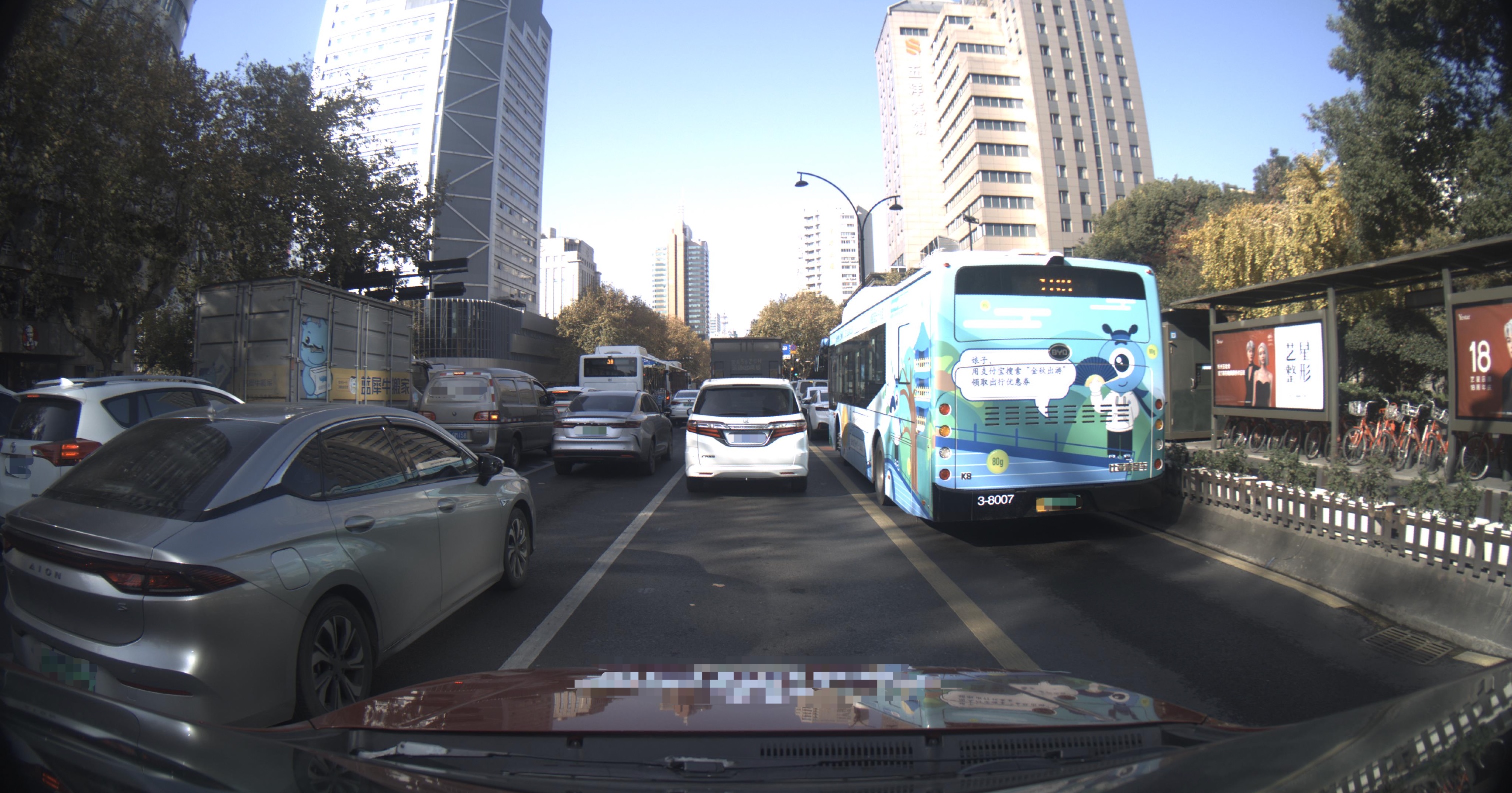}
    \end{minipage}
    \begin{minipage}{0.19\textwidth}
        \centering
        \includegraphics[width=\linewidth]{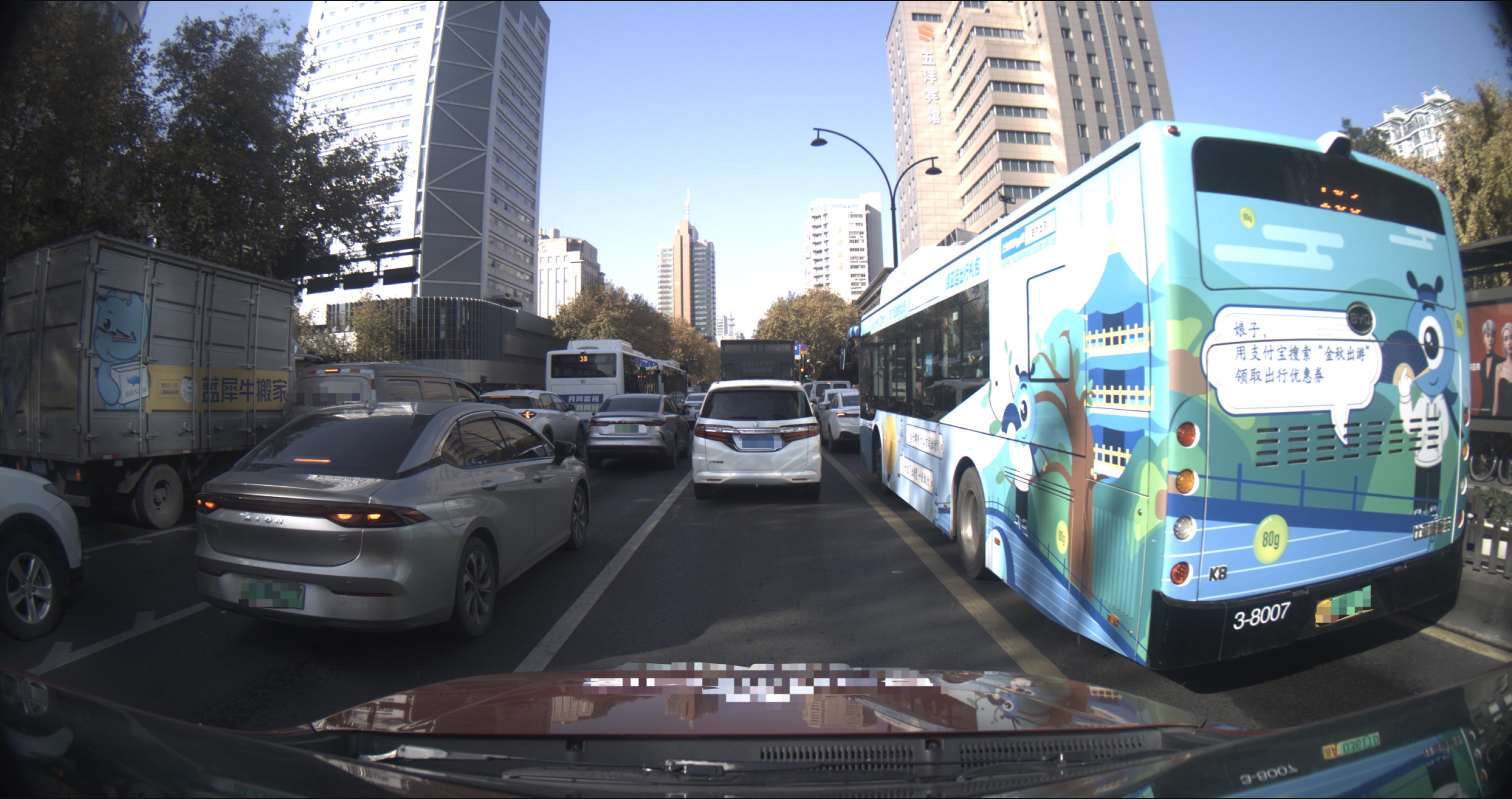}
    \end{minipage}
    \begin{minipage}{0.19\textwidth}
        \centering
        \includegraphics[width=\linewidth]{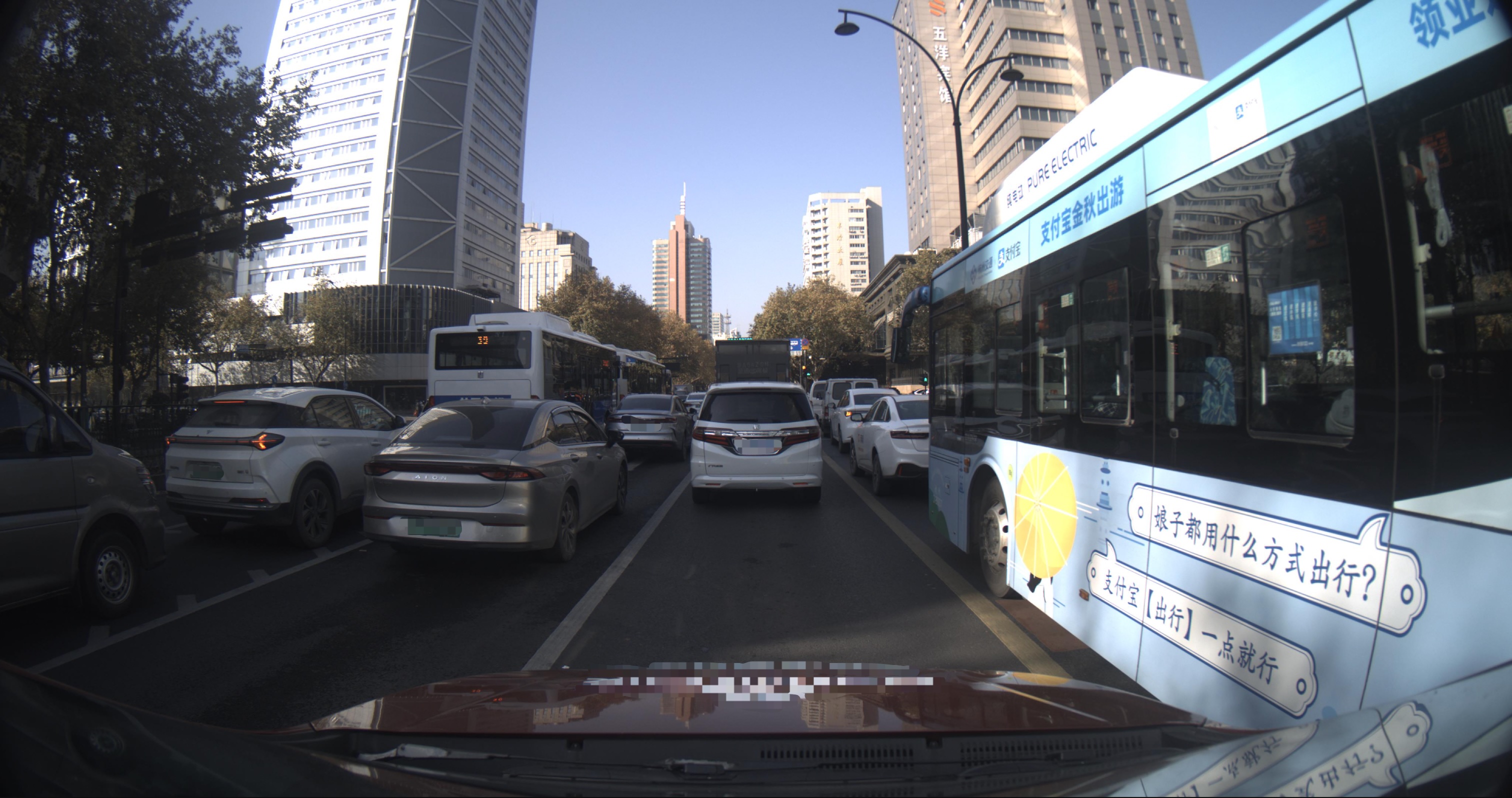}
    \end{minipage}
    \begin{minipage}{0.19\textwidth}
        \centering
        \includegraphics[width=\linewidth]{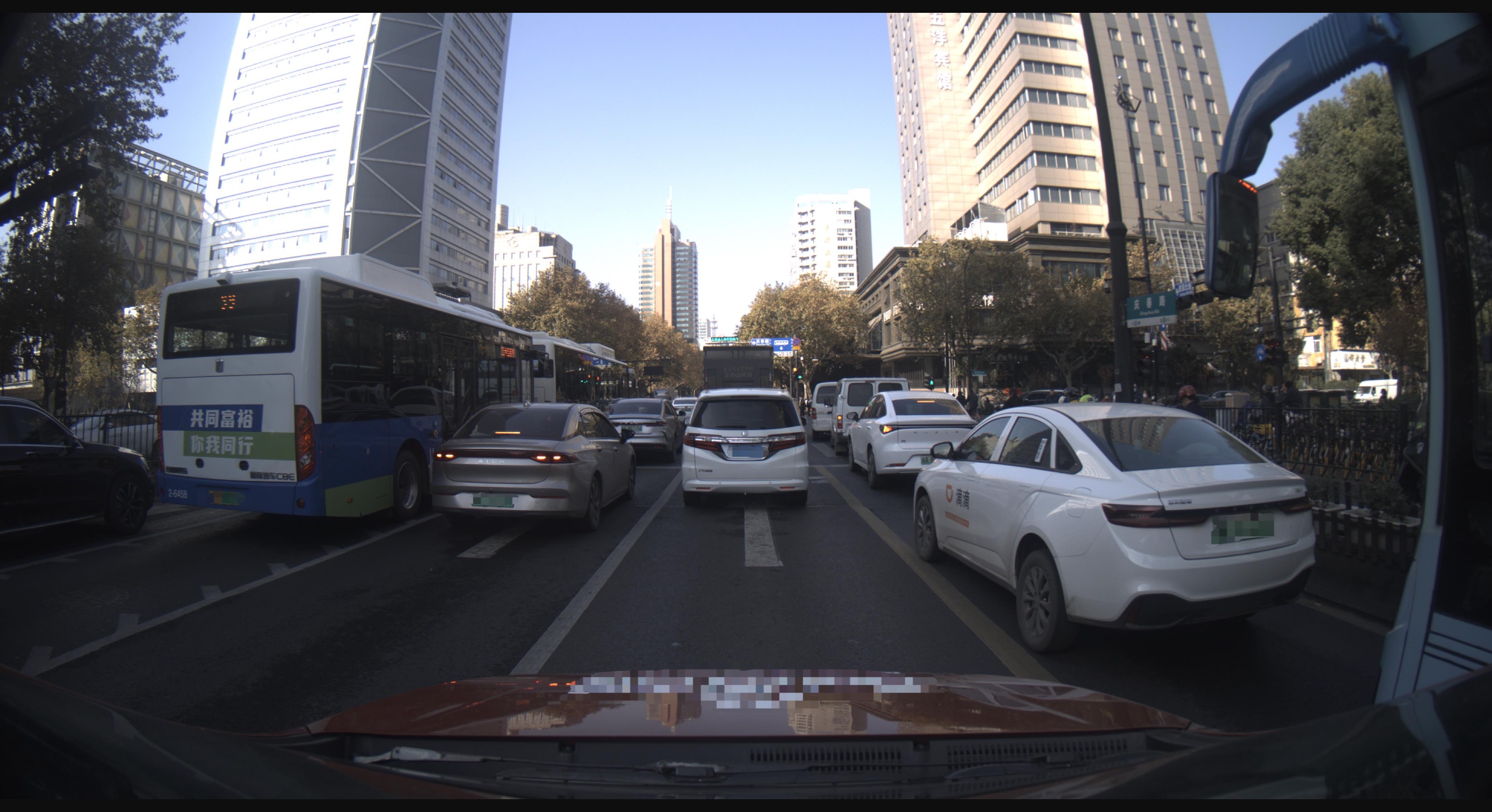}
    \end{minipage}
    \begin{minipage}{0.19\textwidth}
        \centering
        \includegraphics[width=\linewidth]{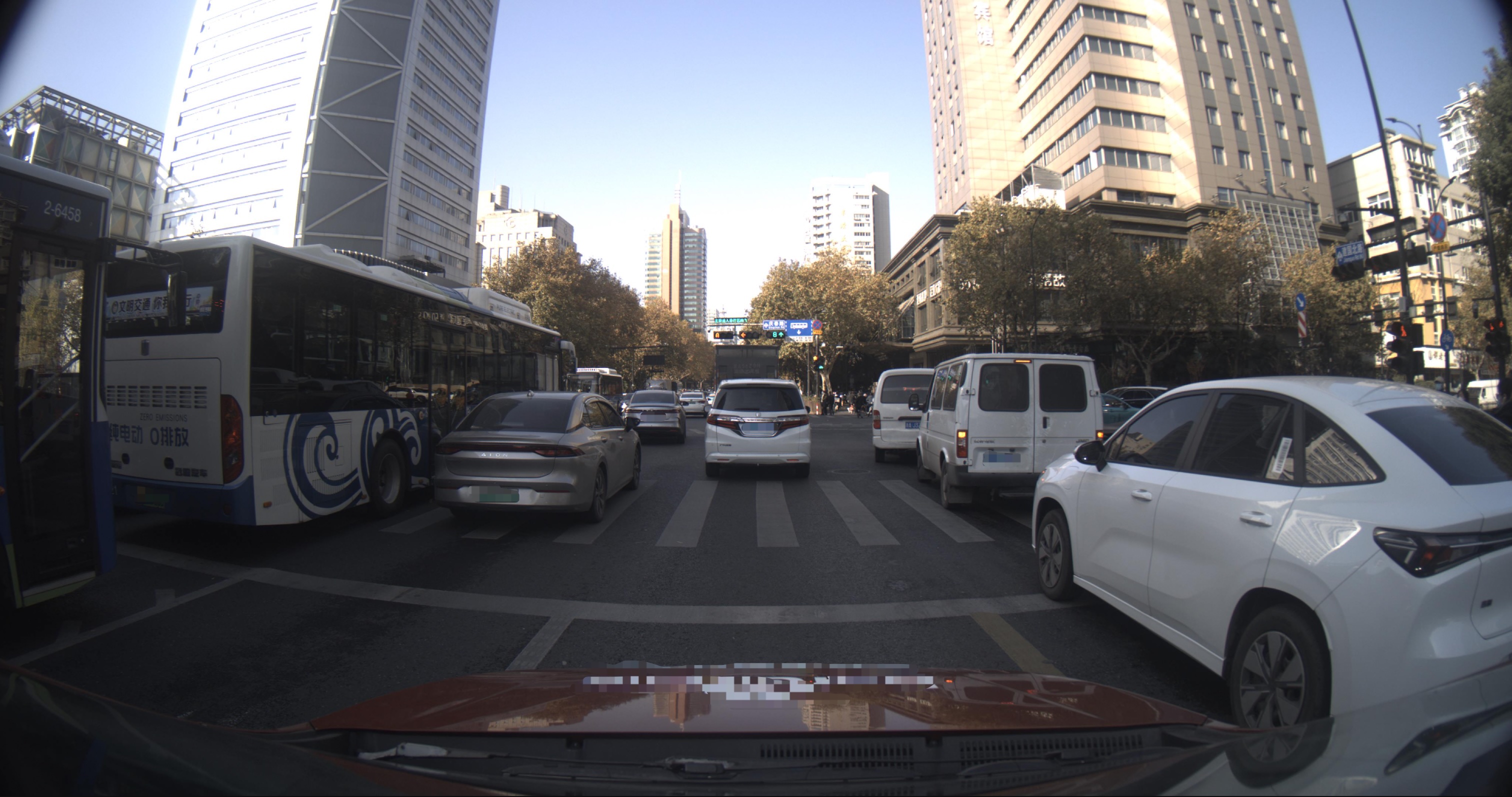}
    \end{minipage}

    \vspace*{-\baselineskip}
    \vspace*{-\parskip}
    \vspace*{3pt} 

    \includegraphics[width=\textwidth, trim=0 0 0 0, clip]{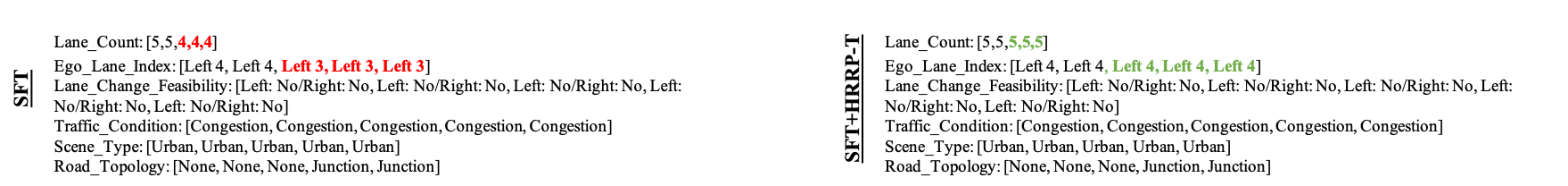}
    \vspace*{-6mm}
    \caption{
    Comparison of SFT and SFT+HRRP-T on a 5-frame congested urban scene. 
    The ego vehicle stays in the same lane: the first two frames clearly show a five-lane layout, whereas the last three frames are partially occluded. 
    SFT reacts to these ambiguous observations with frame-wise drift in lane count and ego-lane index. 
    SFT+HRRP-T leverages temporal evidence and preserves consistent ego-lane predictions with a coherent five-lane topology.
    }
    \label{fig:hrrp-t}
    \vspace{-4mm}
\end{figure*}

\noindent\textbf{Diversity and Challenges of RoadSceneBench.}
First, these examples illustrate the rich diversity of scenarios included in our dataset. The benchmark deliberately samples from complex situations that are critical for autonomous driving. For instance, the cases include: (1) Complex Intersections: \textit{e.g.}, Column 1, Column 3 and Collumn 4, (2)  Congested Traffic Conditions: \textit{e.g.}, Column 2, \textit{Traffic\_Condition: Congested}, (3) Challenging Lighting: \textit{e.g.}, strong shadows and high contrast in Column 2 and Column 3, (4) Ambiguous Lane Change Scenarios: \textit{e.g.}, the presence of a solid line on both left and right in Column 2. The inclusion of these varied and demanding scenarios highlights the significant value of {RoadSceneBench} for robustly evaluating VLM reasoning.

\noindent\textbf{Comparative Model Performance.}
Second, these cases reveal the limitations of current leading VLMs and the superiority of our {MapVLM}. Both the advanced closed-source model, {Gemini-2.5-pro}, and the capable open-source model, {InternVL3-78B}, make several critical errors.
{In contrast, our MapVLM consistently provides predictions that are highly aligned with the GT across all four challenging cases.} This qualitative evidence not only validates the difficulty of our benchmark but also strongly underscores the advanced situational awareness and reasoning capabilities of our method.

As illustrated in Figure~\ref{fig:hrrp-t}, HRRP-T is designed to enforce temporally consistent structural predictions across multi-frame clips in the presence of occlusion and local ambiguity. Compared with the SFT baseline, which reacts myopically to frame-wise appearance changes and produces inconsistent lane-count and ego-lane estimates, SFT+HRRP-T leverages temporal evidence and the no–lane-change prior to maintain a stable ego-lane index and a coherent lane topology at the scene level. These qualitative behaviors are consistent with the aggregate results: compared with SFT, SFT+HRRP-T increases overall precision and recall to 75.78\% and 72.17\%, respectively, with a substantial improvement on Ego-lane Index (69.34\% $\rightarrow$ 75.44\% accuracy, 50.37\% $\rightarrow$ 84.67\% recall). Together, the qualitative and quantitative findings indicate that HRRP-T enhances scene-level reliability across interdependent mid-level tasks, even when the per-frame lane-count precision changes only mildly.

%% file: sec/5_conclusion.tex
\section{Conclusion}
\label{sec:conclusion}
We presented RoadSceneBench, a compact benchmark for mid-level road scene reasoning, and introduced HRRP-T, a temporally consistent reward framework that enhances structural and semantic alignment in VLMs. Results show clear gains in geometry-aware and relational understanding across diverse road scenarios. We believe RoadSceneBench and HRRP-T form a solid foundation for future research on multi-frame reasoning and map-centric scene understanding. In future, we plan to extend the benchmark to broader geographic regions and richer relational semantics, including dynamic road events such as short-term construction, accidents, and temporary lane closures. We also aim to incorporate object grounding and interaction-level reasoning to support finer-grained, map-centric scene understanding.

%% file: sec/X_suppl.tex
\clearpage
\onecolumn
\appendix
\section{Appendix}

\begin{center}
\includegraphics[width=\linewidth]{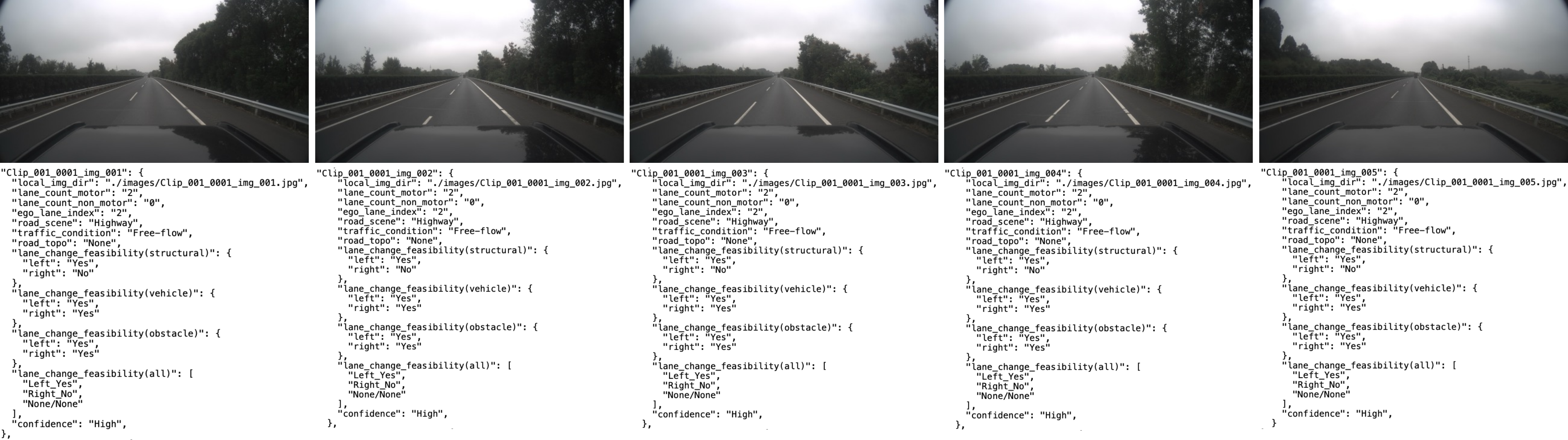}
\includegraphics[width=\linewidth]{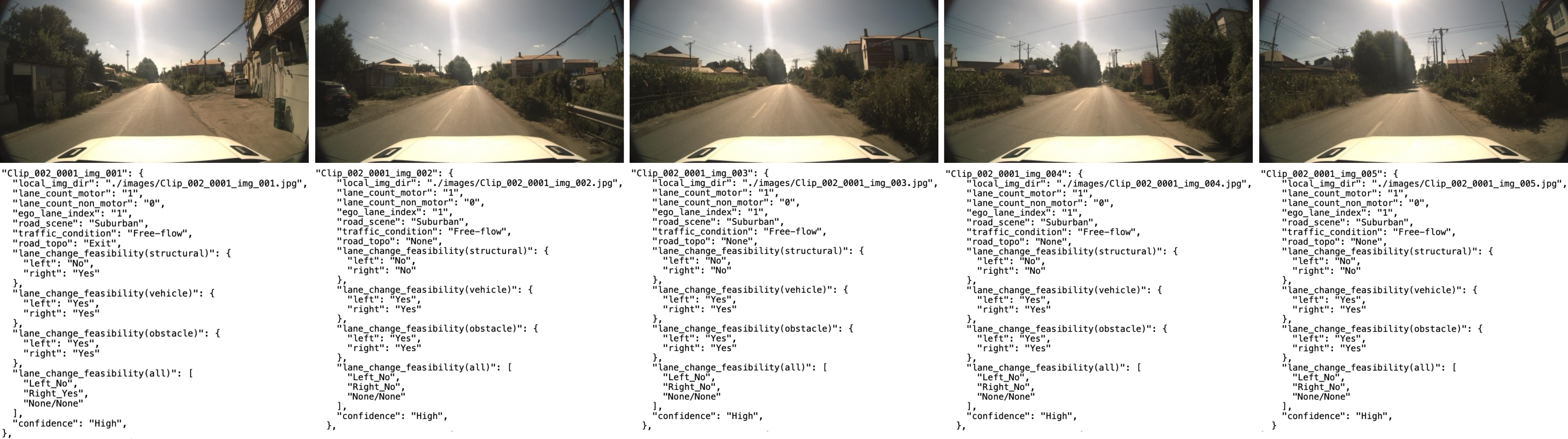}
\includegraphics[width=\linewidth]{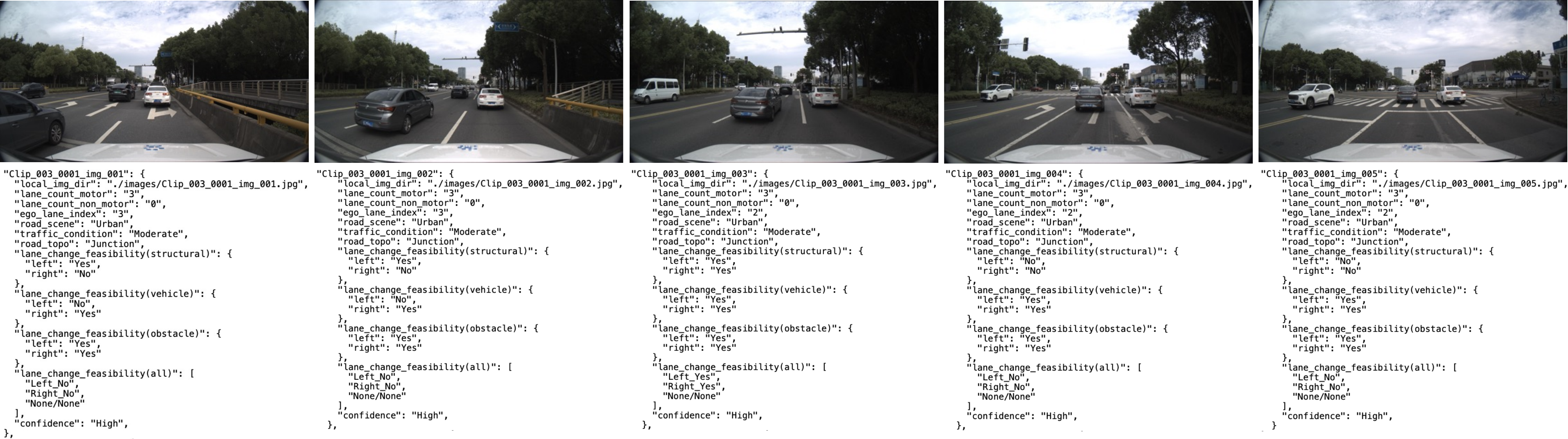}
\includegraphics[width=\linewidth]{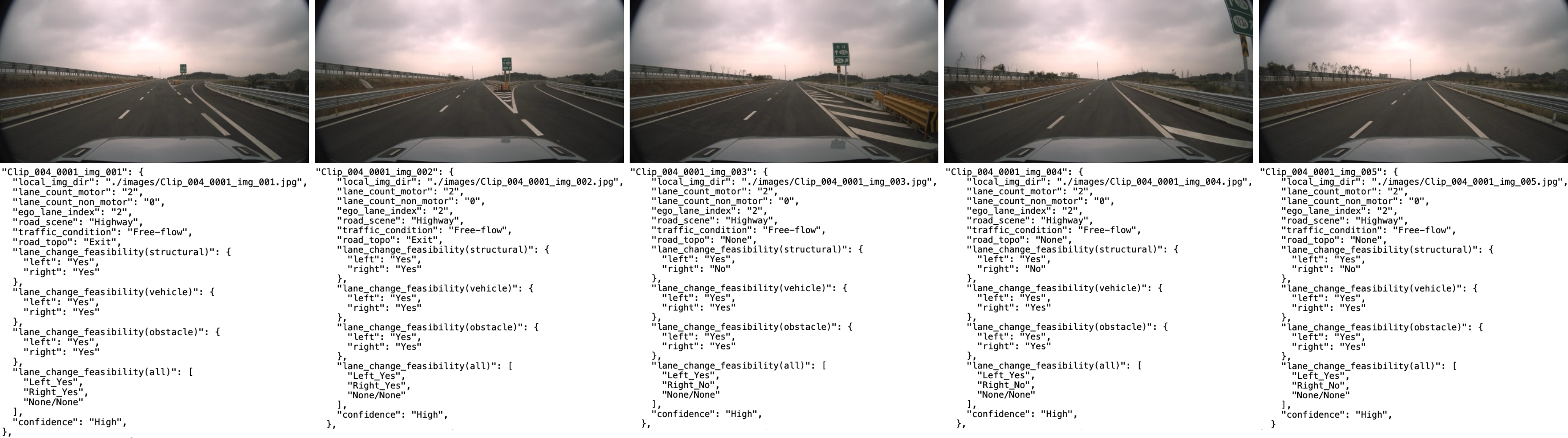}
\includegraphics[width=\linewidth]{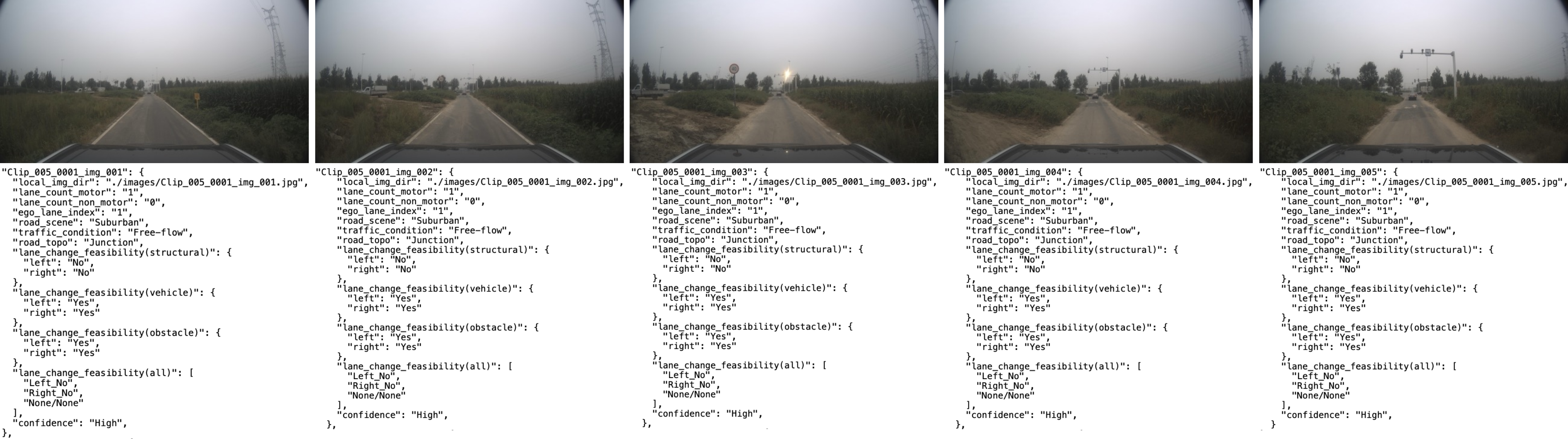}
\includegraphics[width=\linewidth]{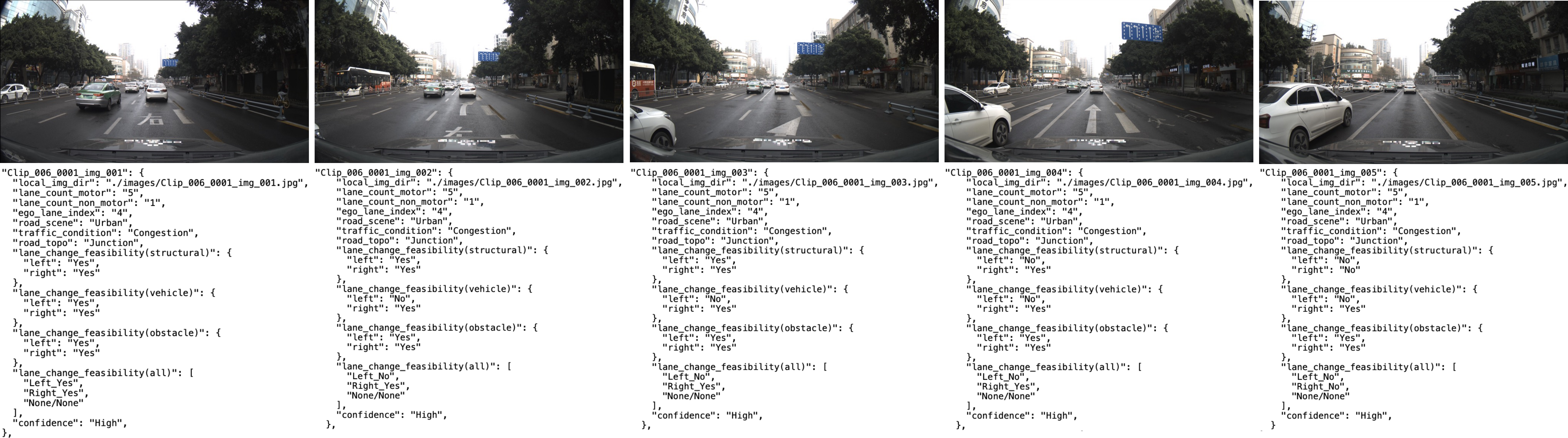}
\captionof{figure}{Examples of RoadSceneBench. Each row displays the images contained in a clip along with their corresponding annotation information.}
\label{fig:exp_1}
\end{center}